\documentclass{article}

\usepackage[nonatbib,final]{neurips_2025}

\usepackage[utf8]{inputenc} 
\usepackage[T1]{fontenc}    
\usepackage{hyperref}       
\usepackage{url}            
\usepackage{bm}
\usepackage{booktabs}       
\usepackage{amsfonts}       
\usepackage{nicefrac}       
\usepackage{microtype}      
\usepackage{xcolor}         
\usepackage{multirow}
\usepackage{subcaption}
\usepackage{graphicx}
\usepackage{amsmath}
\usepackage{cleveref}
\usepackage{tcolorbox}
\usepackage{colortbl}
\usepackage{array}
\usepackage{wrapfig}

\usepackage[backend=biber,
            style=numeric,
            sorting=nyt,         
            maxbibnames=99,
            natbib=true,
            doi=false,
            isbn=false,
            url=false,
            eprint=true]{biblatex}

\DeclareFieldFormat
  [article,inbook,incollection,inproceedings,patent,thesis,
   techreport,unpublished,misc,book]
  {title}{\mkbibquote{#1}}

\newtcolorbox{callout}{
  colback=blue!5,      
  colframe=blue!40,    
  left=3pt, right=3pt, top=3pt, bottom=3pt,
  boxrule=0.5pt,
  arc=2pt
}

\newcommand{\atk}{\texttt{vec2vec}}

\newcommand{\R}{\mathbb{R}}

\newcommand{\shortpara}[1]{\vspace{.75ex}\noindent\textbf{#1}}

\title{Harnessing the Universal Geometry of Embeddings}

\author{%
  Rishi Jha \quad Collin Zhang \quad Vitaly Shmatikov \quad John X. Morris\\
  Department of Computer Science\\
  Cornell University
}

\begin{document}

\maketitle

\begin{abstract}
We introduce the first method for translating text embeddings from one vector space to another without any paired data, encoders, or predefined sets of matches.  Our unsupervised approach translates any embedding to and from a universal latent representation (i.e., a universal semantic structure conjectured by the Platonic Representation Hypothesis).  Our translations achieve high cosine similarity across model pairs with different architectures, parameter counts, and training datasets.

The ability to translate unknown embeddings into a different space while preserving their geometry has serious implications for security.  An adversary with access to a database of only embedding vectors can extract sensitive information about underlying documents, sufficient for classification and attribute inference.

\end{abstract}

\begin{figure}[h]
    \centering
    \includegraphics[width=0.8\linewidth]{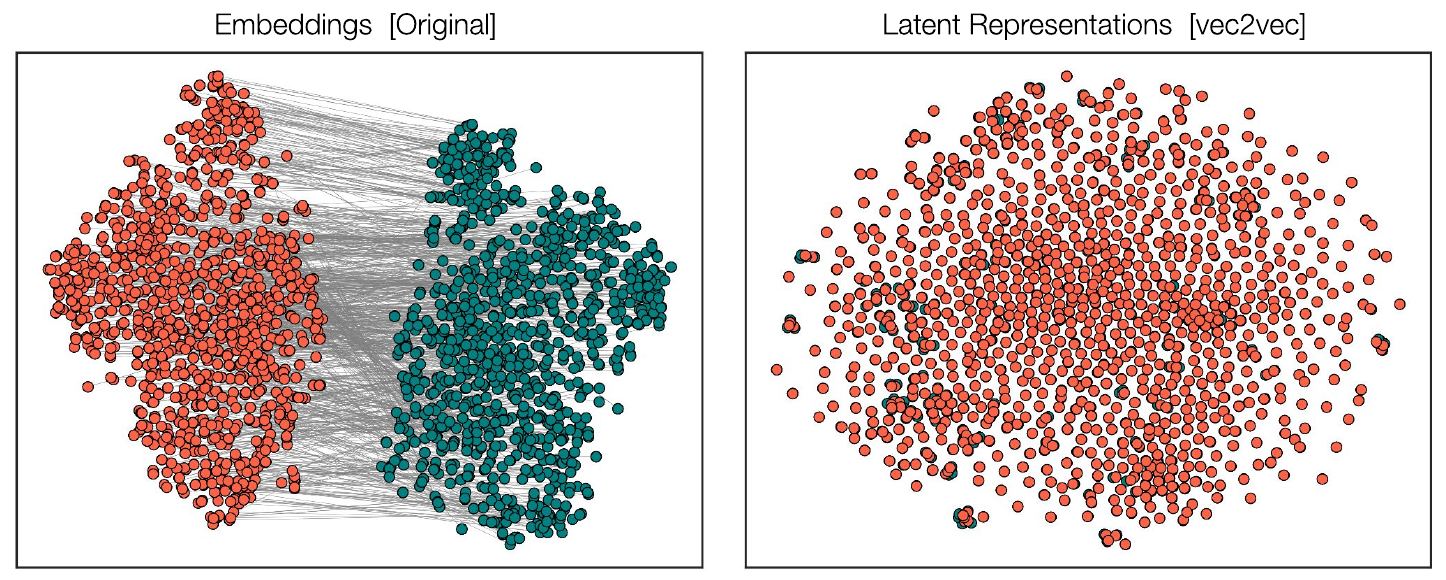}
    \caption{Left: input embeddings from different model families (T5-based GTR \citep{ni2021gtr} and BERT-based GTE \citep{li2023gte}) are fundamentally incomparable. Right:
    given unpaired embedding samples from different models on different texts, our model learns a latent representation where they are closely aligned.}
    \label{fig:main}
\end{figure}

\begin{figure}[t]
    \centering
    \includegraphics[width=0.85\linewidth]{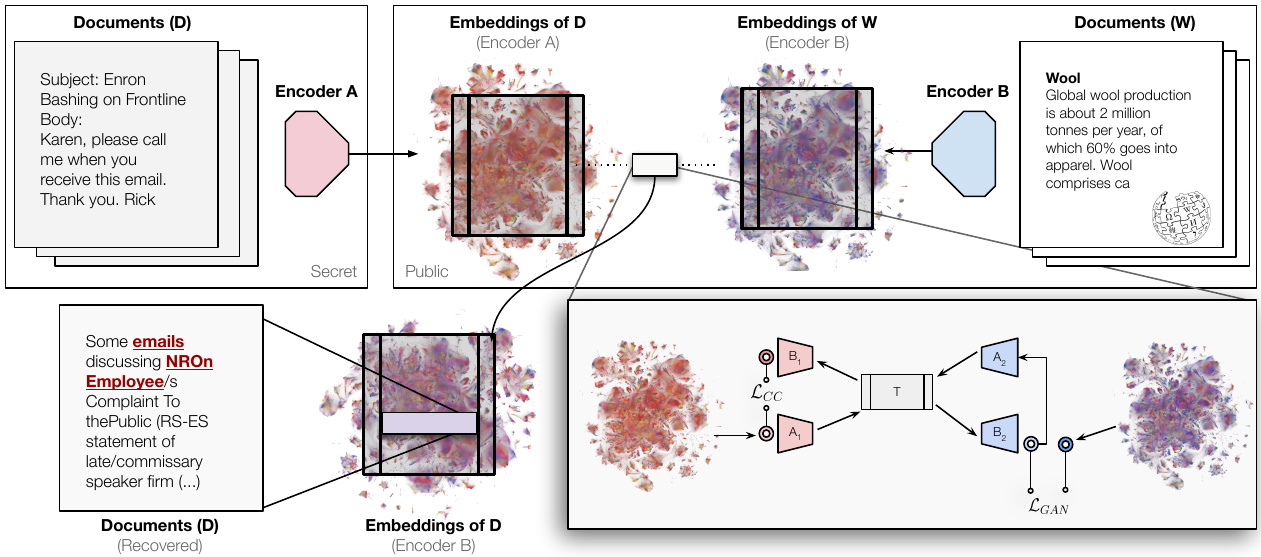}
    \caption{Given only a vector database from an unknown model, \atk{} translates the database into the space of a known model using latent structure alone. Converted embeddings reveal sensitive information about the original documents, such as the topic of an email (pictured, real example).}
    \label{fig:architecture}
\end{figure}
\section{Introduction}

Text embeddings are the backbone of modern NLP, powering tasks like retrieval, RAG, classification, and clustering.  There are many embedding models trained on different datasets, data shufflings, and initializations.  An embedding of a text encodes its semantics: a good model maps texts with similar semantics to vectors close to each other in the embedding space.    Since semantics is a property of text, different embeddings of the same text should encode the same semantics.  In practice, however, different models encode texts into completely different and incompatible vector spaces.




The Platonic Representation Hypothesis \citet{huh2024platonicrepresentationhypothesis} conjectures that all vision models of sufficient size converge to the same latent representation.  We propose a stronger, constructive version of this hypothesis for text models: the universal latent structure of text representations can be learned and, furthermore, harnessed to translate representations from one space to another without any paired data or encoders.

In this work, we show that the Strong Platonic Representation Hypothesis holds in practice.  Given unpaired examples of embeddings from two models with different architectures and training data, our method learns a latent representation in which the embeddings are almost identical (\Cref{fig:main}).

We draw inspiration from research on aligning word embeddings across languages \citep{xing2015normalized, conneau2018wordtranslationparalleldata, grave2018unsupervisedalignmentembeddingswasserstein, chen2018unsupervisedmultilingualwordembeddings} and unsupervised image translation
\citep{liu2018unsupervisedimagetoimagetranslationnetworks,zhu2020unpairedimagetoimagetranslationusing}. Our \atk{} method uses adversarial losses and cycle consistency to learn to encode embeddings into a shared latent space and decode with minimal loss.  This makes unsupervised translation possible.  We use a basic adversarial approach with vector space preservation \citep{mrksic2016counterfittingwordvectorslinguistic} to learn a mapping from an unknown embedding distribution to a known one.

\atk{} is the
\textbf{first method to successfully translate embeddings from the space of one model to another without paired data.}\footnote{Prior work has successfully translated \emph{word} embeddings between languages, typically relying on overlapping vocabularies across languages. In contrast, we translate embeddings of entire sequences between model spaces.}
\atk{}
translations achieve cosine similarity as high as $0.96$ to the ground-truth vectors in their target embedding spaces and perfect matching on over $8000$ shuffled embeddings (without access to the set of possible matches in advance). 

To show that our translations preserve not only the relative geometry of embeddings but also the semantics of underlying inputs, we extract information from them using zero-shot attribute inference and inversion, without any knowledge of the model that produced the original embeddings.\footnote{Our code is available \href{https://github.com/rjha18/vec2vec/}{on GitHub.}}



\section{Problem formulation: unsupervised embedding translation}

Consider a collection of embedding vectors $\{u_1, \ldots u_n\}$, for example, a dump of a compromised vector database, where each $u_i = M_1(d_i)$ is generated by an unknown encoder $M_1 : \mathbb{V}^s \rightarrow \R^{d_{M_1}}$ from an unknown document $d_i$.  We cannot make queries to $M_1$ and do not know its training data, nor architectural details.  Our goal is to extract any information about the documents $d_i$.


\begin{figure}[t]
    \centering
    \includegraphics[width=0.9\linewidth]{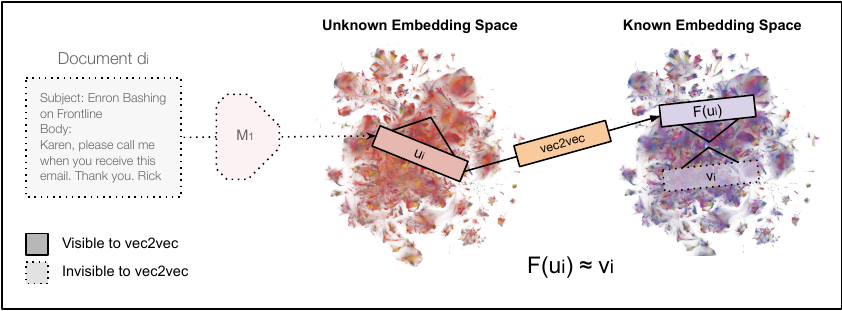}
    \caption{\textbf{Unsupervised embedding translation}. With access to only $u_i = M_1(d_i)$, \atk{} seeks to generate a translation $F(u_i)$ that is close in $M_2$'s embedding space to the ideal embedding $v_i = M_2(d_i)$ without access to $d_i$, $v_i$, or $M_1$.}
    \label{fig:spaces}
\end{figure}

We do assume access to a different encoder $M_2$ that we can query at will to generate new embeddings in some other space.  We also assume high-level distributional knowledge about the hidden documents: their modality (text) and language (e.g., English).  To extract information, we may translate $\{u_1, \ldots u_n\}$ into the output space of $M_2$ and apply techniques like inversion that require the encoder.

\shortpara{Limitations of correspondence methods.}
There is significant prior research on the problem of \textit{matching} or \textit{correspondence} between sets of embedding vectors \citep{alvarezmelis2018gromovwassersteinalignmentwordembedding, peyre2016gromov, chen2020graphoptimaltransportcrossdomain, schnaus2025itsblindmatchvisionlanguage}. These methods typically assume that the two (or more) sets of embeddings are generated by different encoders on the \emph{same or highly-overlapping inputs}. In other words, for each unknown vector, there must already exist a set of candidate vectors in a different embedding. In practice, it is unrealistic to expect that such a database be available, so these methods are not directly applicable. Some matching methods, however, support translation between embedding spaces without overlapping inputs. Our experiments demonstrate that these methods struggle significantly, even when correspondence exists.

Our task is inherently more challenging than matching, because we do not assume access to encoder $M_1$, nor do we have additional representations of documents ${d_1, \ldots, d_n}$ beyond their embeddings $u_i=M_1(d_i)$. Therefore, we rely solely on unsupervised \textit{translation} from $M_1$ to $M_2$. The effectiveness of such unsupervised translation approaches thus critically depends on identifying and leveraging shared geometric structures within the embedding spaces produced by $M_1$ and $M_2$.

\shortpara{The Strong Platonic Representation Hypothesis.} 
Our hope that unsupervised embedding translation is possible at all rests on the stronger version of the Platonic Representation Hypothesis \citep{huh2024platonicrepresentationhypothesis}.  Our conjecture is as follows:
\emph{neural networks trained with the same objective and modality, but with different data and model architectures, converge to a universal latent space such that a translation between their respective representations can be learned without any pairwise correspondence.} 


\shortpara{Translation enables information extraction.} Solving unsupervised translation will allow us to use information extraction tools designed to operate on vectors produced by known encoders. For example, we could apply inversion models \citep{morris2023textembeddingsrevealalmost, zhang2025universalzeroshotembeddinginversion} to recover unknown documents $\{d_i\}$.


\section{Our method: \atk{}}
\label{sec:architecture}

Unsupervised translation has been successful in computer vision, using a combination of cycle consistency and adversarial regularization \citep{liu2018unsupervisedimagetoimagetranslationnetworks,zhu2020unpairedimagetoimagetranslationusing}.  Our design of \atk{} is inspired in part by these methods.  We aim to learn embedding-space translations that are cycle-consistent (mapping to and from an embedding space should end in the same place) and indistinguishable (embeddings for the same text from either space should have identical latents). 


\subsection{Architecture} 
\label{subsec:method-arch}

We propose a modular architecture, where embeddings are encoded and decoded using space-specific adapter modules and passed through a shared backbone network.   \Cref{fig:architecture} shows these components. {Input adapters} $A_1 :\R^{d} \to \mathbb{R}^{Z}$ and $A_2: \R^{d} \to \mathbb{R}^{Z}$ transform embeddings from each encoder-specific space into a universal latent representation of dimension $Z$. The {shared backbone} $T : \mathbb{R}^{Z} \to \mathbb{R}^{Z}$ extracts a common latent embedding from adapted inputs. Output adapters $B_1 : \mathbb{R}^{Z} \to \R^{d}$ and $B_2: \mathbb{R}^{Z} \to \R^{d}$ translate these common latent embeddings back into the encoder-specific spaces. Thus, translation functions $F_1, F_2$ and additional reconstruction mappings $R_1, R_2$ are defined as:
\begin{equation*}
    F_1=B_2 \circ T \circ A_1,\quad F_2=B_1 \circ T \circ A_2\quad R_1=B_1 \circ T \circ A_1\quad R_2=B_2 \circ T \circ A_2
\end{equation*}
Parameters of all components are collectively denoted $\theta = \{A_1, A_2, T, B_1, B_2\}$.


Unlike images, embeddings do not have any spatial bias.  Instead of CNNs, we use multilayer perceptrons (MLP) with residual connections, layer normalization, and SiLU nonlinearities.  Discriminators mirror this structure but omit residual connections to simplify adversarial learning.


\subsection{Optimization}
\label{subsec:method-opt}

In addition to the `generator' networks $F$ and $R$, we introduce discriminators operating on both the latent representations of $F$ ($D_{1}^{\ell}, D_{2}^{\ell}$) and the output embeddings ($D_1, D_2$).

Our goal is to train the parameters of $\theta$ by solving:
\begin{equation}\label{eq:base_objective}
    \theta^* = \arg \min_{\theta} \max_{D_1, D_2, D_{1}^{\ell}, D_{2}^{\ell}} \mathcal{L}_{\text{adv}}(F_1,F_2,D_1,D_2,D_{1}^{\ell},D_{2}^{\ell}) + \lambda_{\text{gen}} \mathcal{L}_{\text{gen}}(\theta),
\end{equation}
where $\mathcal{L}_{\text{adv}}$ and $\mathcal{L}_{\text{gen}}$ represent adversarial and generator-specific constraints respectively and hyperparameter $\lambda_{\rm gen}$ controls their tradeoff.

\shortpara{Adversarial.}
The adversarial loss encourages generated embeddings to match the empirical distributions of original embeddings both at the embedding and latent levels. Specifically, applying the standard GAN loss formulation \citep{10.1145/3422622} to both levels yields:
\begin{align*}
    \mathcal{L}_{\text{adv}}(F_1,F_2,D_1,D_2,D_{1}^{\ell},D_{2}^{\ell}) 
    &= \mathcal{L}_{\text{GAN}}(D_1, F_1) + \mathcal{L}_{\text{GAN}}(D_2, F_2) \\
    &\quad + \mathcal{L}_{\text{GAN}}(D_{1}^{\ell}, T \circ A_1) + \mathcal{L}_{\text{GAN}}(D_{2}^{\ell},  T \circ A_2).
\end{align*}

\shortpara{Generator.}
Because adversarial losses alone do not guarantee that translated embeddings preserve semantics \citep{zhu2020unpairedimagetoimagetranslationusing}, we introduce three additional constraints to help the generator learn a useful mapping:

\textit{Reconstruction} enforces that an embedding, when mapped into the latent space and back into its original embedding space, closely matches its initial representation:
\begin{align*}
    \mathcal{L}_{\text{rec}}(R_1,R_2) = \mathbb{E}_{x \sim p}\|R_1(x)-x\|_2^2 + \mathbb{E}_{y \sim q}\|R_2(y)-y\|_2^2.
\end{align*}

where $p$ and $q$ are distributions of embeddings sampled from $M_1$ and $M_2$, respectively.

\textit{Cycle-consistency} acts as an unsupervised proxy for supervised pair alignment, ensuring that $F$ and $G$ can translate an embedding to the other embedding space and back again with minimal corruption:
\begin{align*}
    \mathcal{L}_{\text{CC}}(F_1,F_2) = \mathbb{E}_{x \sim p}\|F_2(F_1(x))-x\|_2^2 + \mathbb{E}_{y \sim q}\|F_1(F_2(y))-y\|_2^2.
\end{align*}
\textit{Vector space preservation (VSP)} ensures that pairwise relationships between translated embeddings are consistent with the target space \citep{mrksic2016counterfittingwordvectorslinguistic, yoon2025embeddingconverter}. Given a batch of $B$ embeddings $x_1, ..., x_B$ and $y_1, ..., y_B$, we sum their average pairwise distances after translation by both $F_1$ and $F_2$:
\begin{align*}
    \mathcal{L}_{\text{VSP}}(F_1,F_2) = \frac{1}{B^2} \sum_{i=1}^{B} \sum_{j=1}^{B} \bigg[ & \|M_1(x_i) \cdot M_1(x_j) - F_2(M_2(y_i)) \cdot F_2(M_2(y_j))\|_2^2 \\
    & +  \|M_2(y_i) \cdot M_2(y_j) - F_1(M_1(x_i)) \cdot F_1(M_1(x_j))\|_2^2 \bigg]
\end{align*}

Combining these losses yields:
$\mathcal{L}_{\text{gen}}(\theta) = \lambda_{\text{rec}}\mathcal{L}_{\text{rec}}(R_1,R_2) + \lambda_{\text{CC}}\mathcal{L}_{\text{CC}}(F_1,F_2) + \lambda_{\text{VSP}}\mathcal{L}_{\text{VSP}}(F_1,F_2)$, where hyperparameters $\lambda_{\text{CC}}$, $\lambda_{\text{rec}}$, and $\lambda_{\text{VSP}}$ control relative importance.


\section{Experimental setup}
\label{sec:setup}

\subsection{Preliminaries}
\label{sec:prelims}

\shortpara{Datasets.} 
We use the \textit{Natural Questions (NQ)} \cite{kwiatkowski-etal-2019-natural} dataset of user queries and Wikipedia-sourced answers for training (a $2$-million subset) and evaluation (a $65536$ subset).  To evaluate information extraction, we use
\textit{TweetTopic} \cite{antypas-etal-2024-multilingual}, a dataset of tweets multi-labeled by 19 topics; a random $8192$-record subset of
\textit{Pseudo Re-identified MIMIC-III (MIMIC)} \cite{lehman-etal-2021-bert}, a pseudo re-identified version of the MIMIC dataset~\cite{johnson2016mimic} of patient records multi-labeled by 2673 MedCAT \cite{kraljevic2021multi} disease descriptions; and a random $50$-email subset of
the \textit{Enron Email Corpus (Enron)} \cite{10.1007/978-3-540-30115-8_22}, an unlabeled, public dataset of
internal emails from a defunct energy company.  In \Cref{sec:mscoco}, we ablate a model on \textit{MS COCO} \cite{lin2014microsoft}, a captioned image dataset, to evaluate performance on multimodal retrieval.

\shortpara{Models.} \Cref{tab:model-specs} lists the embedding models representing four size categories, five transformer backbones, and two output dimensionalities.  Granite is multilingual; CLIP is multimodal. Since Qwen is very compute-intensive, we only evaluate it for a single model pair in \Cref{sec:qwen}.

\begin{table}[ht]
  \scriptsize
  \centering
  \begin{tabular}{lclccc}
    \toprule
    Model    & Params (M) & Backbone & Year & Dims & Max Seq. \\
    \midrule
    \citep{ni2021gtr} \ gtr      & 110     & T5 & 2021 & 768 & 512  \\
    \citep{radford2021clip} \ clip    & 151     & CLIP & 2021 & 512 & 77  \\
    \citep{wang2024e5} \ e5     & 109     & BERT & 2022 & 768 & 512 \\
    \citep{li2023gte} \ gte    & 109     & BERT & 2023 & 768 & 512 \\
    \citep{zhang2025stella} \ stella & 109     & BERT & 2023 & 768 & 512 \\
    \citep{granite2024embedding} \ granite  & 278     & RoBERTa & 2024 & 768 & 512 \\
    \cite{qwen3embedding} \ qwen & 4000 & Qwen3 & 2025 & 2560 & 32K\\
    \bottomrule\\[-1.6ex]
  \end{tabular}
  \caption{Embedding models used in our experiments.}
  \label{tab:model-specs}
\end{table}

\shortpara{Training.}
Unless otherwise specified, each \atk{} is trained on two sets of embeddings generated from disjoint sets of $1$ million $64$-token sequences sampled from NQ (see \Cref{sec:ablations} for experiments with fewer embeddings).  Due to GAN instability \cite{10.1145/3446374}, we select the best of multiple initializations (see \Cref{sec:stability}) and leave more robust training to future work. See \Cref{sec:compute} for compute details.

\subsection{Evaluating translation}
\label{sec:ot}

Let $u_i = M_1(d_i)$ and $v_i = M_2(d_i)$ denote the source and target embeddings of the same input $d_i$.  The goal of translation is to generate a vector that is as close to $v_i$ as possible.  We say that \((u_i, v_j)\) are ``aligned'' by the translator \(F\) if \(v_j\) is the closest embedding to \(F(u_i)\): $j = \arg\min_k \cos\bigl(F(u_i), v_k\bigr).$
A perfect translator \(F^*\) satisfies \(i = \arg\min_k \cos\bigl(F^*(u_i), v_k\bigr)\) for all \(i\).


Given (unknown) embeddings $\{M_2(d_j)\}_{j=0}^n$ ordered by decreasing cosine similarity to $F(u_i)$, let $r_i$ be the rank of the correct embedding $v_i=M_2(d_i)$.  To measure quality of $F$, we use three metrics. \textbf{Mean Cosine Similarity} measures how close translations are, on average, to their targets. \textbf{Top-1 Accuracy} is the fraction of translations whose target is closer than any other embedding. \textbf{Mean Rank} is the average
rank of targets with respect to translations. The ideal translator \(F^*\) achieves mean similarity of \(1.0\), top-1 accuracy of \(1.0\), and mean rank of \(1.0\). Recall that a random alignment corresponds to a mean rank of $\frac{n}{2}$. Formally,
\begin{equation*}
    \cos(u_i, v_i)= \frac{1}{n}\sum_{i=1}^{n}\bigl[1 - \cos\bigl(F(u_i), v_i\bigr)\bigr] \quad \text{Top-1}(r)=\frac{1}{n}\sum_{i=1}^{n}\mathbf{1}\{r_i = 1\}\quad \text{Rank(r)}=\frac{1}{n}\sum_{i=1}^{n} r_i
\end{equation*}

\atk{} is the first unsupervised embedding translator, thus there is no direct baseline. As our \textbf{Naïve} baseline, we simply use $F(x)=x$ to measure geometric similarity between embedding spaces.  The second (pseudo)baseline is \textbf{Oracle-aided optimal transport}.  It assumes that candidate targets are known and is thus strictly easier than \atk{} and the Naïve baseline.  We solve optimal assignment, $\pi^* = \arg\min_\pi \sum_{i=1}^n \cos(u_i, v_{\pi(i)})$, via either the Hungarian, Earth Mover’s Distance, Sinkhorn, or (Entropic) Gromov-Wasserstein algorithms, choosing the solver with the lowest rank for each experiment. See \Cref{sec:ot_baseline} for more details.

\subsection{Evaluating information extraction}\label{sec:extracting_attributes}

We measure whether translation preserves semantics via \textit{attribute inference}:
for each translated embedding $F(M_1(d_i))$, our goal is to infer attributes $c_i \subseteq \mathcal{C}$ of $d_i$.

The first method we use is \textbf{zero-shot embedding attribute inference}: calculate pairwise cosine similarities between $F(M_1(d_i))$ and the embeddings of all attributes in $\mathcal{C}$, identify top $k$ closest attributes, and measure whether they are correct via \textit{top-$k$ accuracy}: $\frac{1}{n}\sum_{i=0}^{n}\mathbf{1}\left\{|c^k_i\cap c_i| \geq 1\right\}$.

The second method is \textbf{embedding inversion} that recovers text inputs from embeddings.  Since \cite{morris2023textembeddingsrevealalmost} requires a pre-trained inversion model for each embedding space, we use \cite{zhang2025universalzeroshotembeddinginversion} instead to generate an approximation $d'_i$ of $d_i$ from $F(M_1(d_i))$ in a zero-shot manner.  We measure the extracted information using \textit{LLM judge accuracy}: the fraction of translated embeddings for which GPT-4o determines that $d'$ reveals information in $d$. See \Cref{sec:prompt} for our prompt. 

In addition to the Naïve baseline, we also consider an \textbf{Oracle attribute inference}: zero-shot classification with the correct embedding $M_2(d)$ and class labels $M_2(\mathcal{C})$.
 
\section{\atk{} learns to translate embeddings without any paired data}
\label{sec:eval_geometry}



We first show that \atk{} learns a universal latent space, then demonstrate that this space preserves the geometry of all embeddings.  Therefore, we can use it like a \textbf{universal language of text encoders} to translate their representations without any paired data.

\begin{figure}[t]
    \centering
    \includegraphics[width=\linewidth]{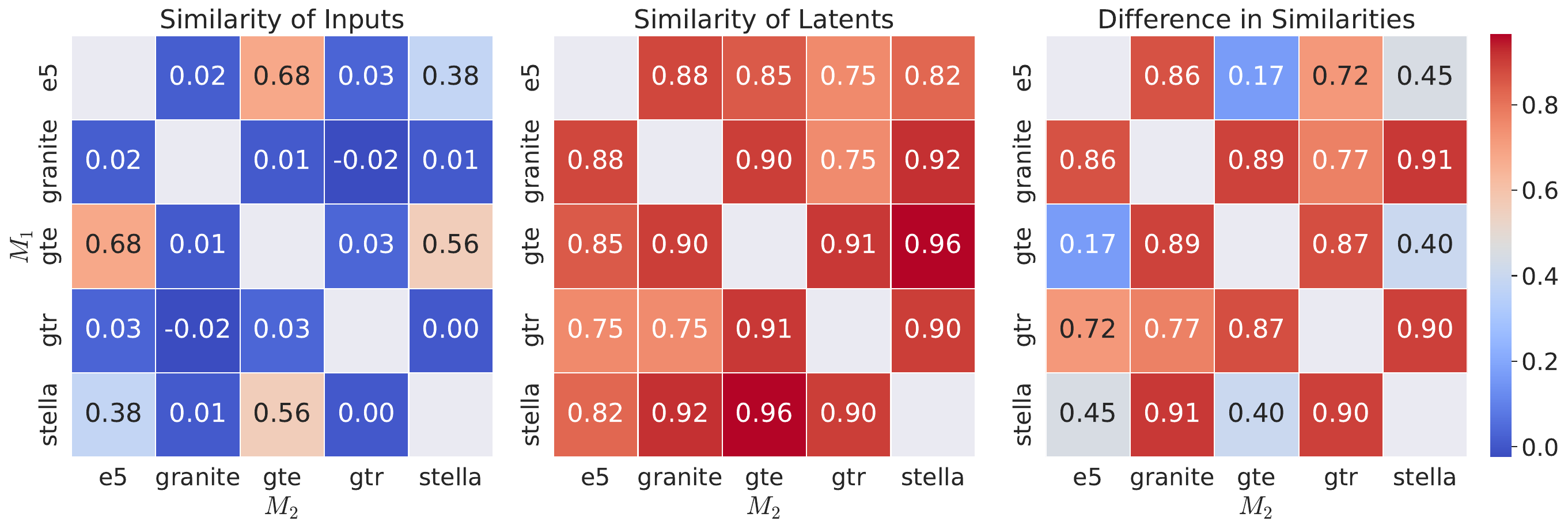}
    \caption{Pairwise cosine similarities of input embeddings (left) and their \atk{} latents (middle) across different embedding pairs. The absolute difference between the heatmaps plots is on the right. All numbers are computed on the same batch of 1024 NQ texts.}
    \label{fig:cosine_heatmaps}
\end{figure}

\shortpara{\atk{} learns a universal latent space.}
\atk{} projects embeddings $M_{1,2,\ldots}$ into a shared latent space via compositions of input adapters ($A_{1,2,\ldots}$) and a shared translator $T$.  \Cref{fig:cosine_heatmaps} shows that even when the embeddings $u_i = M_1(d_i)$ and $v_i = M_2(d_i)$ are far apart (\textit{i.e.,} have low cosine similarity), their representations in \atk{}'s latent space are incredibly close: $T(A_1(u_i)) \approx T(A_2(v_i))$.  \Cref{fig:main} visualizes this (via two-dimensional projections) for \atk{} trained on GTE and GTR embeddings: the embeddings are far apart, but their latents 
are \emph{nearly overlapping}.

\begin{table}[!b]
    \scriptsize
    \centering
    \begin{tabular}{llrrrrrrrrr}
        \toprule
        \multicolumn{2}{c}{} & \multicolumn{3}{c}{\atk{}} & \multicolumn{3}{c}{Naïve Baseline} & \multicolumn{3}{c}{OT Baseline}\\
        \cmidrule(lr){3-5} \cmidrule(lr){6-8}  \cmidrule(lr){9-11} 
        $M_1$ & $M_2$ & $\cos(\cdot)$ $\uparrow$ & T-1 $\uparrow$ & Rank $\downarrow$ & $\cos(\cdot)$ $\uparrow$ & T-1 $\uparrow$ & Rank $\downarrow$ &  $\cos(\cdot)$ $\uparrow$ & T-1 $\uparrow$ & Rank $\downarrow$\\
        \midrule
        \multirow{4}{*}{gra.} & gtr & \textbf{0.80 {\tiny (0.0)}} & \textbf{0.99} & \textbf{1.19 {\tiny (0.1)}} & -0.03 {\tiny (0.0)} & 0.00 & 4168.73 {\tiny (9.2)} & 0.70 {\tiny (0.0)} & 0.00 & 2773.72 {\tiny (8.6)}$^{\ddagger}$ \\
        & gte & \textbf{0.87 {\tiny (0.0)}} & \textbf{0.95} & \textbf{1.18 {\tiny (0.0)}} & 0.01 {\tiny (0.0)} & 0.00 & 4088.58 {\tiny (9.2)} & 0.85 {\tiny (0.0)} & 0.00 & 2680.02 {\tiny (8.6)}$^{\ddagger}$ \\
        & ste. & \textbf{0.79 {\tiny (0.0)}} & \textbf{0.98} & \textbf{1.05 {\tiny (0.0)}} & 0.01 {\tiny (0.0)} & 0.00 & 4208.26 {\tiny (9.2)} & 0.67 {\tiny (0.0)} & 0.00 & 3446.52 {\tiny (8.8)}$^{\ddagger}$ \\
        & e5 & \textbf{0.85 {\tiny (0.0)}} & \textbf{0.98} & \textbf{1.11 {\tiny (0.0)}} & 0.02 {\tiny (0.0)} & 0.00 & 4111.60 {\tiny (9.2)} & 0.83 {\tiny (0.0)} & 0.00 & 3569.59 {\tiny (8.7)}$^{\ddagger}$ \\
        \midrule
        \multirow{4}{*}{gtr} & gra. & \textbf{0.81 {\tiny (0.0)}} & \textbf{0.99} & \textbf{1.02 {\tiny (0.0)}} & -0.03 {\tiny (0.0)} & 0.00 & 4169.76 {\tiny (9.2)} & 0.70 {\tiny (0.0)} & 0.00 & 2775.17 {\tiny (8.6)}$^{\ddagger}$ \\
        & gte & \textbf{0.87 {\tiny (0.0)}} & \textbf{0.93} & \textbf{2.31 {\tiny (0.1)}} & 0.04 {\tiny (0.0)} & 0.00 & 4080.92 {\tiny (9.2)} & 0.85 {\tiny (0.0)} & 0.00 & 3070.69 {\tiny (8.9)}$^{\ddagger}$ \\
        & ste. & \textbf{0.80 {\tiny (0.0)}} & \textbf{0.99} & \textbf{1.03 {\tiny (0.0)}} & 0.00 {\tiny (0.0)} & 0.00 & 4198.78 {\tiny (9.2)} & 0.67 {\tiny (0.0)} & 0.00 & 3559.06 {\tiny (9.1)}$^{\ddagger}$ \\
        & e5 & \textbf{0.83 {\tiny (0.0)}} & \textbf{0.84} & \textbf{2.88 {\tiny (0.2)}} & 0.03 {\tiny (0.0)} & 0.00 & 4082.84 {\tiny (9.2)} & \textbf{0.83 {\tiny (0.0)}} & 0.00 & 3888.01 {\tiny (8.9)}$^{\ddagger}$ \\
        \midrule
        \multirow{4}{*}{gte} & gra. & \textbf{0.75 {\tiny (0.0)}} & \textbf{0.95} & \textbf{1.22 {\tiny (0.0)}} & 0.01 {\tiny (0.0)} & 0.00 & 4079.81 {\tiny (9.3)} & 0.69 {\tiny (0.0)} & 0.00 & 2664.38 {\tiny (8.6)}$^{\ddagger}$ \\
        & gtr & \textbf{0.75 {\tiny (0.0)}} & \textbf{0.91} & \textbf{2.64 {\tiny (0.1)}} & 0.04 {\tiny (0.0)} & 0.00 & 4084.15 {\tiny (9.2)} & 0.70 {\tiny (0.0)} & 0.00 & 3064.16 {\tiny (8.9)}$^{\ddagger}$ \\
        & ste. & \textbf{0.89 {\tiny (0.0)}} & \textbf{1.00} & \textbf{1.00 {\tiny (0.0)}} & 0.56 {\tiny (0.0)} & \textbf{1.00} & \textbf{1.00 {\tiny (0.0)}} & 0.71 {\tiny (0.0)} & \textbf{1.00} & \textbf{1.00 {\tiny (0.0)}$^{\dagger}$} \\
        & e5 & \textbf{0.87 {\tiny (0.0)}} & 0.99 & 5.19 {\tiny (0.5)} & 0.68 {\tiny (0.0)} & \textbf{1.00} & \textbf{1.00 {\tiny (0.0)}} & 0.84 {\tiny (0.0)} & \textbf{1.00} & \textbf{1.00 {\tiny (0.0)}$^{\dagger}$} \\
        \midrule
        \multirow{4}{*}{ste.} & gra. & \textbf{0.80 {\tiny (0.0)}} & \textbf{0.98} & \textbf{1.08 {\tiny (0.0)}} & 0.01 {\tiny (0.0)} & 0.00 & 4209.08 {\tiny (9.3)} & 0.69 {\tiny (0.0)} & 0.00 & 3419.44 {\tiny (8.8)}$^{\ddagger}$ \\
        & gtr & \textbf{0.82 {\tiny (0.0)}} & \textbf{1.00} & \textbf{1.10 {\tiny (0.0)}} & 0.00 {\tiny (0.0)} & 0.00 & 4192.31 {\tiny (9.2)} & 0.70 {\tiny (0.0)} & 0.00 & 3555.64 {\tiny (9.0)}$^{\ddagger}$ \\
        & gte & \textbf{0.92 {\tiny (0.0)}} & \textbf{1.00} & \textbf{1.00 {\tiny (0.0)}} & 0.56 {\tiny (0.0)} & \textbf{1.00} & \textbf{1.00 {\tiny (0.0)}} & 0.87 {\tiny (0.0)} & \textbf{1.00} & \textbf{1.00 {\tiny (0.0)}$^{\dagger}$} \\
        & e5 & \textbf{0.86 {\tiny (0.0)}} & \textbf{1.00} & \textbf{1.00 {\tiny (0.0)}} & 0.38 {\tiny (0.0)} & 0.99 & 1.03 {\tiny (0.0)} & 0.83 {\tiny (0.0)} & \textbf{1.00} & \textbf{1.00 {\tiny (0.0)}$^{\dagger}$} \\
        \midrule
        \multirow{4}{*}{e5} & gra. & \textbf{0.81 {\tiny (0.0)}} & \textbf{0.99} & \textbf{2.20 {\tiny (0.2)}} & 0.02 {\tiny (0.0)} & 0.00 & 4120.60 {\tiny (9.3)} & 0.69 {\tiny (0.0)} & 0.00 & 3526.02 {\tiny (8.7)}$^{\ddagger}$ \\
        & gtr & \textbf{0.74 {\tiny (0.0)}} & \textbf{0.82} & \textbf{2.56 {\tiny (0.0)}} & 0.03 {\tiny (0.0)} & 0.00 & 4080.76 {\tiny (9.3)} & 0.70 {\tiny (0.0)} & 0.00 & 3877.03 {\tiny (8.8)}$^{\ddagger}$ \\
        & gte & \textbf{0.90 {\tiny (0.0)}} & \textbf{1.00} & 1.01 {\tiny (0.0)} & 0.68 {\tiny (0.0)} & \textbf{1.00} & \textbf{1.00 {\tiny (0.0)}} & 0.86 {\tiny (0.0)} & \textbf{1.00} & \textbf{1.00 {\tiny (0.0)}$^{\dagger}$} \\
        & ste. & \textbf{0.78 {\tiny (0.0)}} & \textbf{1.00} & \textbf{1.00 {\tiny (0.0)}} & 0.38 {\tiny (0.0)} & \textbf{1.00} & \textbf{1.00 {\tiny (0.0)}} & 0.69 {\tiny (0.0)} & \textbf{1.00} & \textbf{1.00 {\tiny (0.0)}$^{\dagger}$} \\
        \bottomrule\\[-1.6ex]
    \end{tabular}
    \caption{In-distribution translations: \atk{}s trained on NQ and evaluated on a 65536 text subset of NQ (chunked in batches of size 8192). The rank metric varies from 1 to 8192, thus 4096 corresponds to a random ordering. Standard errors are shown in parentheses. Bold denotes best value. Symbols denote the lowest-rank solver for specific experiments: Sinkhorn$^\dagger$ and Gromov-Wasserstein$^\ddagger$.}
    \label{tab:embedding_comparison}
\end{table}

\shortpara{\atk{} translations mirror target geometry.}
\Cref{tab:embedding_comparison} shows that \atk{} generates embeddings with near-optimal assignment across model pairs, achieving cosine similarity scores up to 0.92, top-1 accuracies up to 100\%, and ranks as low as 1. In same-backbone pairings (e.g., (gte, e5)), \atk{}'s top-1 accuracy and rank are comparable to both the naïve baseline and (surprisingly) the oracle-aided optimal transport. Although the embeddings generated by \atk{} are significantly closer to the ground truth than the naïve baseline, in same-backbone pairings the embeddings are close enough to be compatible.  In cross-backbone pairings, \atk{} is far superior on all metrics, while baseline methods perform similarly to random guessing.



\begin{table}[t]
    \scriptsize
    \centering
    \begin{tabular}{llrrr|rrr}
        \toprule
        \multicolumn{2}{c}{} & \multicolumn{3}{c}{TweetTopic} & \multicolumn{3}{c}{MIMIC}\\
        \cmidrule(lr){3-5} \cmidrule(lr){6-8}
        $M_1$ & $M_2$ & $\cos(\cdot)$ $\uparrow$& T-1 $\uparrow$& Rank $\downarrow$& $\cos(\cdot)$ $\uparrow$& T-1 $\uparrow$& Rank $\downarrow$\\
        \midrule
        \multirow{4}{*}{gran.} & gtr & 0.74 {\tiny (0.0)} & 0.99 & 1.09 {\tiny (0.1)} & 0.74 {\tiny (0.0)} & 0.60 & 23.38 {\tiny (1.6)} \\
        & gte & 0.85 {\tiny (0.0)} & 0.95 & 1.26 {\tiny (0.1)} & 0.85 {\tiny (0.0)} & 0.08 & 346.21 {\tiny (7.8)} \\
        & stel. & 0.77 {\tiny (0.0)} & 0.96 & 1.11 {\tiny (0.0)} & 0.72 {\tiny (0.0)} & 0.13 & 242.23 {\tiny (6.1)} \\
        & e5 & 0.83 {\tiny (0.0)} & 0.87 & 3.10 {\tiny (0.7)} & 0.84 {\tiny (0.0)} & 0.12 & 361.06 {\tiny (8.7)} \\
        \midrule
        \multirow{4}{*}{gtr} & gran. & 0.79 {\tiny (0.0)} & 0.98 & 2.41 {\tiny (0.6)} & 0.78 {\tiny (0.0)} & 0.51 & 35.27 {\tiny (1.9)} \\
        & gte & 0.85 {\tiny (0.0)} & 0.96 & 1.29 {\tiny (0.2)} & 0.84 {\tiny (0.0)} & 0.12 & 279.56 {\tiny (6.9)} \\
        & stel. & 0.77 {\tiny (0.0)} & 0.96 & 1.10 {\tiny (0.0)} & 0.72 {\tiny (0.0)} & 0.27 & 127.92 {\tiny (4.4)} \\
        & e5 & 0.80 {\tiny (0.0)} & 0.53 & 13.38 {\tiny (1.2)} & 0.82 {\tiny (0.0)} & 0.01 & 1413.80 {\tiny (18.3)} \\
        \midrule
        \multirow{4}{*}{gte} & gran. & 0.73 {\tiny (0.0)} & 0.94 & 1.33 {\tiny (0.1)} & 0.73 {\tiny (0.0)} & 0.09 & 342.15 {\tiny (7.8)} \\
        & gtr & 0.71 {\tiny (0.0)} & 0.95 & 1.29 {\tiny (0.1)} & 0.69 {\tiny (0.0)} & 0.12 & 256.63 {\tiny (6.4)} \\
        & stel. & 0.86 {\tiny (0.0)} & 1.00 & 1.00 {\tiny (0.0)} & 0.85 {\tiny (0.0)} & 1.00 & 1.00 {\tiny (0.0)} \\
        & e5 & 0.83 {\tiny (0.0)} & 0.91 & 1.57 {\tiny (0.2)} & 0.86 {\tiny (0.0)} & 0.54 & 17.71 {\tiny (0.9)} \\
        \midrule
        \multirow{4}{*}{stel.} & gran. & 0.79 {\tiny (0.0)} & 0.99 & 1.09 {\tiny (0.1)} & 0.77 {\tiny (0.0)} & 0.14 & 221.95 {\tiny (5.9)} \\
        & gtr & 0.77 {\tiny (0.0)} & 1.00 & 1.00 {\tiny (0.0)} & 0.75 {\tiny (0.0)} & 0.56 & 17.70 {\tiny (1.0)} \\
        & gte & 0.90 {\tiny (0.0)} & 1.00 & 1.00 {\tiny (0.0)} & 0.91 {\tiny (0.0)} & 1.00 & 1.00 {\tiny (0.0)} \\
        & e5 & 0.85 {\tiny (0.0)} & 0.98 & 1.05 {\tiny (0.0)} & 0.85 {\tiny (0.0)} & 0.51 & 26.33 {\tiny (1.2)} \\
        \midrule
        \multirow{4}{*}{e5} & gran. & 0.79 {\tiny (0.0)} & 0.98 & 1.08 {\tiny (0.0)} & 0.78 {\tiny (0.0)} & 0.21 & 151.09 {\tiny (4.6)} \\
        & gtr & 0.67 {\tiny (0.0)} & 0.80 & 3.10 {\tiny (0.6)} & 0.66 {\tiny (0.0)} & 0.01 & 1029.64 {\tiny (14.9)} \\
        & gte & 0.87 {\tiny (0.0)} & 0.99 & 1.02 {\tiny (0.0)} & 0.87 {\tiny (0.0)} & 0.60 & 32.59 {\tiny (2.6)} \\
        & stel. & 0.75 {\tiny (0.0)} & 0.98 & 1.06 {\tiny (0.0)} & 0.75 {\tiny (0.0)} & 0.46 & 32.12 {\tiny (1.4)} \\
        \bottomrule\\[-1.6ex]
    \end{tabular}
    \caption{Out-of-distribution translations: \atk{}s trained on NQ and evaluated on the entire TweetTopic test set (800 tweets) and an 8192-record subset of MIMIC. The rank metric varies from 1 to 800 (for TweetTopic) and 8192 (for MIMIC), thus 400 and, respectively, 4096 correspond to a random ordering. Standard errors are shown in parentheses.}
    \label{tab:embeddings_ood}
\end{table}

\Cref{tab:embeddings_ood} shows that this performance extends to out-of-distribution data.  Our \atk{} translators were trained on NQ (drawn from Wikipedia), yet exhibit high cosine similarity, high accuracy, and low rank when evaluated on tweets (which are far more colloquial and use emojis) and medical records (which contain domain-specific jargon unlikely to appear in NQ). In \Cref{sec:full_ood}, we show that baseline methods fail on cross-backbone embedding pairs.

\begin{table}[!b]
    \scriptsize
    \centering
    \begin{tabular}{llrrrrrr}
        \toprule
        \multicolumn{2}{c}{} & \multicolumn{3}{c}{\atk{}} & \multicolumn{3}{c}{OT Baseline}\\
        \cmidrule(lr){3-5} \cmidrule(lr){6-8}
        $M_1$ & $M_2$ & $\cos(\cdot)$ $\uparrow$ & T-1 $\uparrow$ & Rank $\downarrow$ & $\cos(\cdot)$ $\uparrow$ & T-1 $\uparrow$ & Rank $\downarrow$\\
        \midrule
        gra. & \multirow{5.5}{*}{clip} & \textbf{0.78 {\tiny (0.0)}} & \textbf{0.35} & \textbf{226.62 {\tiny (3.2)}} & 0.76 {\tiny (0.0)} & 0.00 & 4073.58 {\tiny (9.4)}$^{\ddagger}$ \\
        gtr & & \textbf{0.73 {\tiny (0.0)}} & \textbf{0.13} & \textbf{711.23 {\tiny (5.9)}} & 0.59 {\tiny (0.0)} & 0.00 & 4096.78 {\tiny (9.2)}$^{\ddagger}$ \\
        gte & & 0.62 {\tiny (0.0)} & \textbf{0.00} & \textbf{3233.41 {\tiny (9.8)}} & \textbf{0.76 {\tiny (0.0)}} & 0.00 & 4026.96 {\tiny (9.4)}$^{\ddagger}$\\
        ste. & & \textbf{0.77 {\tiny (0.0)}} & \textbf{0.31} & \textbf{286.69 {\tiny (3.6)}} & 0.76 {\tiny (0.0)} & 0.00 & 3955.71 {\tiny (8.9)}$^{\ddagger}$ \\
        e5 & & 0.64 {\tiny (0.0)} & \textbf{0.01} & \textbf{2568.21 {\tiny (9.4)}} & \textbf{0.77 {\tiny (0.0)}} & 0.00 & 3771.52 {\tiny (9.1)}$^{\ddagger}$ \\
        \midrule
        \multirow{5.5}{*}{clip} & gra. & \textbf{0.74 {\tiny (0.0)}} & \textbf{0.72} & \textbf{4.46 {\tiny (0.1)}} & 0.69 {\tiny (0.0)} & 0.00 & 4053.11 {\tiny (9.4)}$^{\ddagger}$\\
         & gtr & \textbf{0.67 {\tiny (0.0)}} & \textbf{0.27} & \textbf{155.11 {\tiny (2.1)}} & 0.49 {\tiny (0.0)} & 0.00 & 4096.35 {\tiny (9.2)}$^{\ddagger}$ \\
         & gte & 0.75 {\tiny (0.0)} & \textbf{0.00} & \textbf{2678.90 {\tiny (8.9)}} & \textbf{0.85 {\tiny (0.0)}} & 0.00 & 4025.81 {\tiny (9.3)}$^{\ddagger}$ \\
         & ste. & \textbf{0.72 {\tiny (0.0)}} & \textbf{0.61} & \textbf{22.50 {\tiny (0.5)}} & 0.67 {\tiny (0.0)} & 0.00 & 3951.73 {\tiny (8.9)}$^{\ddagger}$ \\
         & e5 & 0.73 {\tiny (0.0)} & \textbf{0.01} & \textbf{1692.28 {\tiny (8.2)}} & \textbf{0.83 {\tiny (0.0)}} & 0.00 & 3771.38 {\tiny (9.0)}$^{\ddagger}$ \\
        \bottomrule\\[-1.6ex]
    \end{tabular}
    \caption{Translations between unimodal and multimodal (CLIP) embeddings: \atk{}s trained on NQ and evaluated on a 65536 text subset of NQ (chunked in batches of size 8192). Rank varies from 1 to 8192, thus 4096 corresponds to a random ordering. Since the embedding dimensionalities are different, only the Gromov-Wasserstein$^{\ddagger}$ OT baseline is run and the naive baseline does not apply. Bold denotes best value.}
    \label{tab:CLIP_nq}
\end{table} 

Finally, \Cref{tab:CLIP_nq} shows that \atk{} can even translate to and from the space of CLIP, a multimodal embedding model which was trained in part on \emph{image} data.  While the translations are not as strong as in \Cref{tab:embedding_comparison}, \atk{} consistently outperforms the optimal transport baseline. These results show the promise of our method at adapting to new modalities: in particular, the embedding space of CLIP has been successfully connected to other modalities such as heatmaps, audio, and depth charts \citep{girdhar2023imagebind}.




\section{Using \atk{} translations to extract information}
\label{sec:eval_semantics}

In this section, we show that \atk{} translations not only preserve the geometric structure of embeddings but also retain sufficient semantics to enable attribute inference.



\begin{table}[!t]
    \scriptsize
    \centering
    \begin{tabular}{llcccc|cccc}
        \toprule
        \multicolumn{2}{c}{} & \multicolumn{4}{c}{TweetTopic ($k=1$)} & \multicolumn{4}{c}{MIMIC ($k=10$)}\\
        \cmidrule(lr){3-6} \cmidrule(lr){7-10}
        $M_1$ & $M_2$ & \atk{} & Naïve & $M_1$ & $M_2$ & \atk{} & Naïve & $M_1$ & $M_2$ \\
        \midrule
        \multirow{4}{*}{gran.} & gtr & 0.25 & 0.10 & 0.30 & 0.24 & 0.19 & 0.11 & 0.76 & 0.88 \\
        & gte & 0.32 & 0.09 & 0.30 & 0.34 & 0.36 & 0.13 & 0.76 & 1.00 \\
        & stel. & 0.24 & 0.10 & 0.30 & 0.28 & 0.27 & 0.04 & 0.76 & 0.96 \\
        & e5 & 0.31 & 0.18 & 0.30 & 0.31 & 0.19 & 0.20 & 0.76 & 0.97 \\
        \midrule
        \multirow{4}{*}{gtr} & gran. & 0.34 & 0.08 & 0.24 & 0.30 & 0.16 & 0.12 & 0.88 & 0.76 \\
        & gte & 0.33 & 0.13 & 0.24 & 0.34 & 0.28 & 0.05 & 0.88 & 1.00 \\
        & stel. & 0.30 & 0.10 & 0.24 & 0.28 & 0.25 & 0.07 & 0.88 & 0.96 \\
        & e5 & 0.30 & 0.04 & 0.24 & 0.31 & 0.09 & 0.09 & 0.88 & 0.97 \\
        \midrule
        \multirow{4}{*}{gte} & gran. & 0.37 & 0.04 & 0.34 & 0.30 & 0.18 & 0.11 & 1.00 & 0.76 \\
        & gtr & 0.24 & 0.13 & 0.34 & 0.24 & 0.10 & 0.03 & 1.00 & 0.88 \\
        & stel. & 0.31 & 0.20 & 0.34 & 0.28 & 0.68 & 0.83 & 1.00 & 0.96 \\
        & e5 & 0.37 & 0.30 & 0.34 & 0.31 & 0.37 & 0.63 & 1.00 & 0.97 \\
        \midrule
        \multirow{4}{*}{stel.} & gran. & 0.35 & 0.07 & 0.28 & 0.30 & 0.23 & 0.09 & 0.96 & 0.76 \\
        & gtr & 0.26 & 0.13 & 0.28 & 0.24 & 0.22 & 0.09 & 0.96 & 0.88 \\
        & gte & 0.38 & 0.36 & 0.28 & 0.34 & 0.90 & 0.98 & 0.96 & 1.00 \\
        & e5 & 0.35 & 0.34 & 0.28 & 0.31 & 0.38 & 0.46 & 0.96 & 0.97 \\
        \midrule
        \multirow{4}{*}{e5} & gran. & 0.33 & 0.15 & 0.31 & 0.30 & 0.14 & 0.07 & 0.97 & 0.76 \\
        & gtr & 0.26 & 0.22 & 0.31 & 0.24 & 0.11 & 0.04 & 0.97 & 0.88 \\
        & gte & 0.34 & 0.28 & 0.31 & 0.34 & 0.47 & 0.66 & 0.97 & 1.00 \\
        & stel. & 0.26 & 0.16 & 0.31 & 0.28 & 0.36 & 0.40 & 0.97 & 0.96 \\
        \bottomrule\\[-1.6ex]
    \end{tabular}
    \caption{Information leakage via top-$k$ zero-shot attribute inference: \atk{}s trained on NQ and evaluated on the TweetTopic test set (800 tweets) and an 8192-record subset of MIMIC. $M_1$ and $M_2$ represent \textit{ideal} zero-shot inference: attributes and embeddings are encoded using the same model.}
    \label{tab:tweet_topic_mimic_class}
\end{table}

\begin{wrapfigure}{r}{0.42\textwidth}
    \vspace{-5.5pt}
    \centering
    \includegraphics[width=0.37\textwidth]{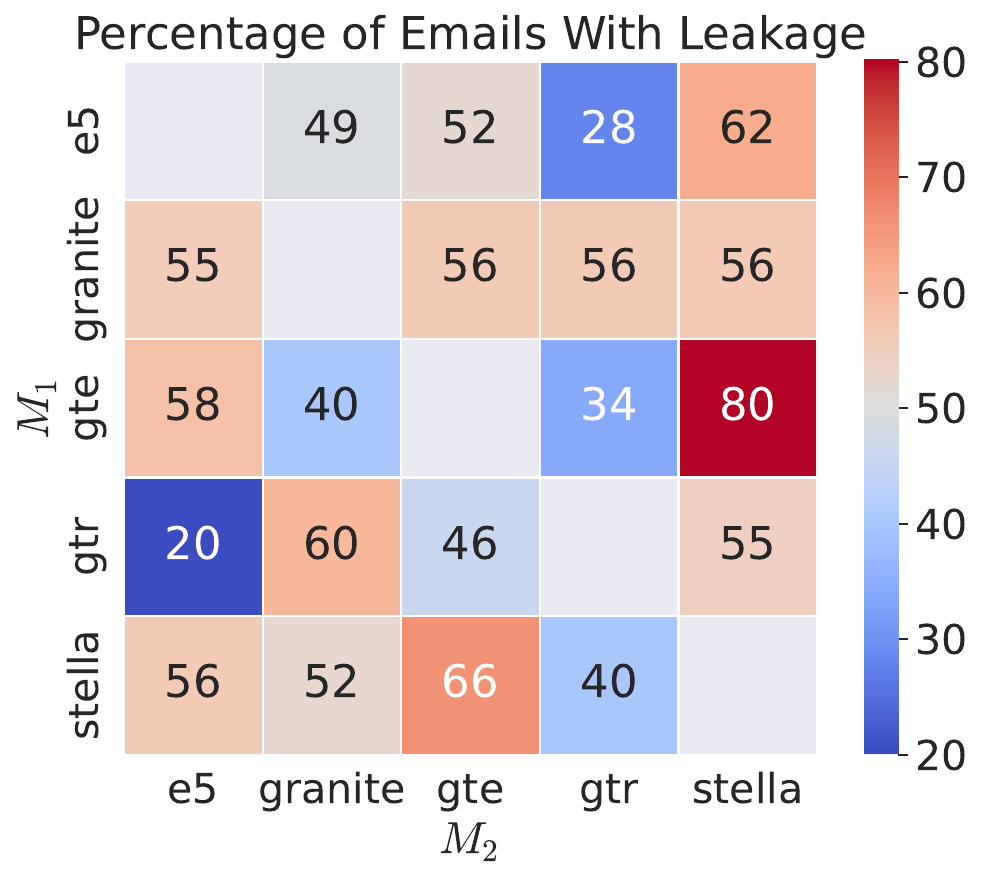}
    \captionsetup{belowskip=-4pt}
    \caption{Leakage of information via inversion. Trained on NQ and evaluated on a 50-email subset of the Enron Email Corpus. Cells denote judge accuracy.}
    \label{fig:enron_heatmap}
\end{wrapfigure}

\shortpara{Zero-shot attribute inference.}
\label{sec:tweet_classification}
\Cref{tab:tweet_topic_mimic_class} shows that attribute inference on \atk{} translations consistently outperforms the naïve baseline and often does better than the ideal zero-shot baseline which performs inference on ground-truth document and attribute embeddings in the same space (this baseline is imaginary since these embeddings are not available in our setting).


\label{sec:mimic_classification}
\atk{} translations even work for embeddings of medical records, which are much further from the training distribution than tweets.  The attributes in this case are MedCAT disease descriptions, very few of which occur in the training data.  Attribute inference on translated embeddings is comparable to the naïve baseline in same-backbone pairings and outperforms it (often greatly) in cross-backbone pairings.  The fact that \atk{} preserves the semantics of concepts like "alveolar periostitis" (which never appears in its training data) is evidence that its latent space is indeed a universal representation.

\begin{figure}[t]
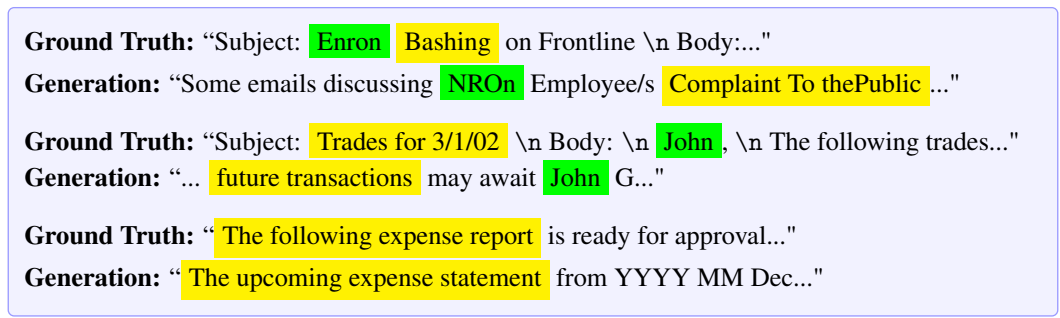

  \centering
    \begin{callout}
    \textbf{Ground Truth:} ``Subject: \colorbox{green}{Enron} \colorbox{yellow}{Bashing} on Frontline \verb|\n| Body:..."\\
    \textbf{Generation:} ``Some emails discussing \colorbox{green}{NROn} Employee/s \colorbox{yellow}{Complaint To thePublic}..."\\
    
    \textbf{Ground Truth:} ``Subject: \colorbox{yellow}{Trades for 3/1/02} \verb|\n| Body: \verb|\n| \colorbox{green}{John}, \verb|\n| The following trades..."\\
    \textbf{Generation:} ``... \colorbox{yellow}{future transactions} may await \colorbox{green}{John} G..."\\

    \textbf{Ground Truth:} ``\colorbox{yellow}{The following expense report} is ready for approval..."\\
    \textbf{Generation:} ``\colorbox{yellow}{The upcoming expense statement} from YYYY MM Dec..."    
    \end{callout}
  \caption{Examples of Enron Email Corpus inversions that infer \colorbox{green}{entities} and \colorbox{yellow}{content}.}
  \label{fig:inversion_example}
\end{figure}

\shortpara{Zero-shot inversion.}
Inversion, i.e., reconstruction of text inputs, is more ambitious than attribute inference.  \atk{} translations retain enough semantic information that off-the-shelf, zero-shot inversion methods like~\cite{zhang2025universalzeroshotembeddinginversion}, developed for embeddings computed by standard encoders, extract information for as many as 80\% of emails and 67\% of tweets given \emph{only} their translated embeddings, for some model pairs (\Cref{fig:enron_heatmap} and \Cref{sec:zs-tt}).  These inversions are imperfect and we leave development of specialized inverters for translated embeddings to future work.  Nevertheless, as exemplified in \Cref{fig:inversion_example}, they still extract potentially sensitive information such as individual and company names, dates, promotions, financial information, outages, and even lunch orders. In \Cref{sec:prompt}, we show the prompt we use to measure extraction.

\section{Ablations}
\label{sec:ablations}

\begin{table}[!ht]
    \scriptsize
    \centering
    \begin{tabular}{lrrr}
        \toprule
        Method & $\cos(\cdot)$ $\uparrow$ & T-1 $\uparrow$ & Rank $\downarrow$\\
        \midrule
        \atk{}&0.75 {\scriptsize (0.0)} & 0.91 & 2.64 {\scriptsize (0.1)}\\
        Naïve Baseline&0.04 {\scriptsize (0.0)} & 0.00 & 4084.15 {\scriptsize (9.2)}\\
        OT Baseline& 0.70 {\scriptsize (0.0)} & 0.00 & 3064.16 {\scriptsize (8.9)}\\
        \midrule
        -- VSP loss&0.58 {\scriptsize (0.0)} & 0.00 & 4196.64 {\scriptsize (9.2)}\\
        -- CC loss&0.50 {\scriptsize (0.0)} & 0.00 & 3941.36 {\scriptsize (9.3)}\\
        -- latent GAN&0.49 {\scriptsize (0.0)} & 0.00 & 3897.09 {\scriptsize (9.5)}\\
        -- VSP \textit{and} CC loss&0.47 {\scriptsize (0.0)} & 0.00 & 3365.24 {\scriptsize (9.3)}\\
        -- hyperparam. tuning&0.50 {\scriptsize (0.0)} & 0.00 & 4011.73 {\scriptsize (9.3)}\\
        \bottomrule\\[-1.6ex]
    \end{tabular}
    \caption{gte $\to$ gtr translators trained without individual components of our method on NQ and evaluated on a 65536-text subset of NQ (chunked in batches of 8192). The rank metric varies from 1 to 8192, thus 4096 corresponds to a random ordering. Standard errors are shown in parentheses.}
    \label{tab:loss_ablation}
\end{table}

\shortpara{Each component of our method is important.} We ablate our method subtractively, measuring the key metrics after removing individual components of our algorithm (described in \Cref{sec:architecture}).  \Cref{tab:loss_ablation} shows that each component appears to be \textit{critical} to building good translations. While \atk{}'s $\cos(\cdot)$ is higher than the naïve baseline, it performs worse across the board than the OT baseline and does not preserve the geometry of the vector space.\\

\begin{table}[!ht]
    \scriptsize
    \centering
    \begin{tabular}{rrrr}
        \toprule
        $N$ & $\cos(\cdot)$ $\uparrow$ & T-1 $\uparrow$ & Rank $\downarrow$ \\
        \midrule
        1000000&0.75 {\scriptsize (0.0)} & 0.92 & 2.73 {\scriptsize (0.2)}\\
        \midrule
        10000 & 0.57 {\scriptsize (0.0)} & 0.01 & 1462.21 {\scriptsize (20.)}\\
        50000 & 0.74 {\scriptsize (0.0)} & 0.81 & 3.91 {\scriptsize (0.6)}\\
        100000  & 0.74 {\scriptsize (0.0)} & 0.85 &  4.52 {\scriptsize (0.4)}\\
        500000 & 0.75 {\scriptsize (0.0)} & 0.92 &  2.73 {\scriptsize (0.2)}\\
        \bottomrule\\[-1.6ex]
    \end{tabular}
    \caption{gte $\to$ gtr translators trained with different amounts of GTE data: \atk{} models trained on NQ and evaluated an 8192-record subset of NQ. The rank metric varies from 1 to 8192, thus 4096 corresponds to a random ordering. Standard errors are shown in parentheses.}
    \label{tab:n_data}
\end{table}

\shortpara{\atk{}s can be trained with significantly less data.} In \Cref{sec:eval_geometry,sec:eval_semantics}, we use 1M-point subsets of NQ to train our \atk{} models. Now, we train the gte $\to$ gtr \atk{} with 1M GTR embeddings but fewer GTE embeddings. \Cref{tab:n_data} shows that with as few as 10K embeddings, the translators still learn something (i.e. are better than random). Translations trained on 50K embeddings are almost as good as those trained on 1M. Translations generally improve with more training data.

\section{Related work}

\shortpara{Representation alignment.}
Similarities between representations of
different neural networks are investigated in \citep{laakso2000content, li2016convergentlearningdifferentneural, wang2018understandinglearningrepresentationsextent, bansal2021revisitingmodelstitchingcompare, huh2024platonicrepresentationhypothesis, wolfram2025layers, li-etal-2024-vision-language}. Methods based on CCA \citep{morcos2018insightsrepresentationalsimilarityneural}, SVCCA, \citep{raghu2017svccasingularvectorcanonical}, CKA \citep{kornblith2019similarity, maniparambil2024vision}, ICA \citep{yamagiwa2023discoveringuniversalgeometryembeddings}, time-series \citep{mistry2023comparative}, and GUIs \citep{heimerl2019embComp} have been used to compare embeddings from different subspaces.  
\citep{maiorca2024latentspacetranslationsemantic, moschella2023relativerepresentationsenablezeroshot, moayeri2023texttoconceptandbackcrossmodel, tian2019latent, norelli2023asifcoupleddataturns} harness representation similarity for zero-shot stitching, substitution, domain transfer, and multimodal adaptation.  All rely on some amount of paired data, which is difficult  
to reduce \citep{cannistraci2023bootstrappingparallelanchorsrelative}.  Our method does not just measure similarity, we learn how to \emph{translate} representations across spaces without any paired data.

\shortpara{Optimal transport.}  The problem of unsupervised optimal transport has been studied for image style transfer \citep{huang2018multimodalunsupervisedimagetoimagetranslation, liu2018unsupervisedimagetoimagetranslationnetworks,zhu2020unpairedimagetoimagetranslationusing}, word translation \citep{xing2015normalized, conneau2018wordtranslationparalleldata, grave2018unsupervisedalignmentembeddingswasserstein, chen2018unsupervisedmultilingualwordembeddings, joulin-etal-2018-loss}, and natural language sequence translation \citep{raviknight2011deciphering, lample2018unsupervisedmachinetranslationusing, alvarezmelis2018gromovwassersteinalignmentwordembedding, artetxe2018unsupervisedneuralmachinetranslation, yang2018unsupervisedneuralmachinetranslation, artetxe-etal-2018-robust}.
Our method builds on these works, which often employ a combination of cycle-consistency and adversarial loss. Importantly, unlike prior word and sequence translation methods, multiple representations of the same input (e.g., heavily overlapping word vocabularies) are unavailable in our setting. \citep{schnaus2025itsblindmatchvisionlanguage} proposes a solver for matching small sets of embeddings between different vision-language models.
Our method goes well beyond matching by taking unknown embeddings and \emph{generating} matching embeddings in the space of another model.

\shortpara{Embedding inversion.} An emerging line of research investigates decoding text from language model embeddings \citep{song2020informationleakageembeddingmodels, li2023sentenceembeddingleaksinformation, morris2023textembeddingsrevealalmost} and outputs \citep{morris2023languagemodelinversion, carlini2024stealingproductionlanguagemodel, zhang2024extractingpromptsinvertingllm}.  
\atk{} helps apply these to unknown embeddings, without an encoder or paired data, by translating them to the space of a known model. 

\shortpara{Bridging modality gaps.} Previous work has noted an inherent ``gap” between image- and text-based models \citep{liang2022mindgapunderstandingmodality} and proposed various ways to unify the modalities \citep{song2023bridgegapmodalitiescomprehensive}. Some approaches feed image embeddings directly into language models \citep{koh2023fromage, wang2023visionllm, dong2024dreamllmsynergisticmultimodalcomprehension, liu2023visualinstructiontuning}, while others generate captions from image embeddings \citep{mokady2021clipcap} or even from text embeddings themselves \citep{morris2023textembeddingsrevealalmost}. \citep{girdhar2023imagebind} introduces a shared embedding space that integrates inputs from multiple modalities, including text, audio, and vision. In contrast, our post-hoc approach directly translates between representations and complements these systems by enabling inputs from a wide variety of embedding models.
\section{Discussion and Future Work}

The Platonic Representation Hypothesis conjectures that the representation spaces of modern neural networks are converging.   We assert the Strong Platonic Representation Hypothesis: the latent universal representation can be learned and harnessed to translate between representation spaces without any encoders or paired data.

In \Cref{sec:eval_geometry}, we demonstrated 
that our \atk{} method successfully translates embeddings generated from unseen documents by unseen encoders, and the translator is robust to (sometimes very) out-of-distribution inputs.  This suggests that \atk{} learns domain-agnostic translations based on the universal geometric relationships which encode the same semantics in multiple embedding spaces.


In \Cref{sec:eval_semantics}, we showed that \atk{} translations preserve sufficient input semantics to enable attribute inference. 
We extracted sensitive disease information from patient records and partial content from corporate emails, with access only to document embeddings and no access to the encoder that produced them.  Better translation methods will enable higher-fidelity extraction, confirming once again that embeddings reveal (almost) as much as their inputs.


Our findings provide compelling evidence for the Strong Platonic Representation Hypothesis for text-based models.  Our preliminary results on CLIP suggest that the universal geometry can be harnessed in other modalities, too.  The results in this paper are but a \emph{lower bound} on inter-representation translation.  Better and more stable learning algorithms, architectures, and other methodological improvements will support scaling to more data, more model families, and more modalities.


\begin{ack}

This research is supported in part by the Google Cyber NYC Institutional Research Program. RJ is supported by the Digital Life Initiative Fellowship and JM by the National Science Foundation.


\end{ack}

\printbibliography

@misc{alvarezmelis2018gromovwassersteinalignmentwordembedding,
      title={Gromov-Wasserstein Alignment of Word Embedding Spaces}, 
      author={David Alvarez-Melis and Tommi S. Jaakkola},
      year={2018},
      eprint={1809.00013},
      archivePrefix={arXiv},
      primaryClass={cs.CL},
      url={https://arxiv.org/abs/1809.00013}, 
}

@misc{artetxe2018unsupervisedneuralmachinetranslation,
      title={Unsupervised Neural Machine Translation}, 
      author={Mikel Artetxe and Gorka Labaka and Eneko Agirre and Kyunghyun Cho},
      year={2018},
      eprint={1710.11041},
      archivePrefix={arXiv},
      primaryClass={cs.CL},
      url={https://arxiv.org/abs/1710.11041}, 
}

@misc{bansal2021revisitingmodelstitchingcompare,
      title={Revisiting Model Stitching to Compare Neural Representations}, 
      author={Yamini Bansal and Preetum Nakkiran and Boaz Barak},
      year={2021},
      eprint={2106.07682},
      archivePrefix={arXiv},
      primaryClass={cs.LG},
      url={https://arxiv.org/abs/2106.07682}, 
}

@misc{cannistraci2023bootstrappingparallelanchorsrelative,
      title={Bootstrapping Parallel Anchors for Relative Representations}, 
      author={Irene Cannistraci and Luca Moschella and Valentino Maiorca and Marco Fumero and Antonio Norelli and Emanuele Rodolà},
      year={2023},
      eprint={2303.00721},
      archivePrefix={arXiv},
      primaryClass={cs.LG},
      url={https://arxiv.org/abs/2303.00721}, 
}

@misc{carlini2024stealingproductionlanguagemodel,
      title={Stealing Part of a Production Language Model}, 
      author={Nicholas Carlini and Daniel Paleka and Krishnamurthy Dj Dvijotham and Thomas Steinke and Jonathan Hayase and A. Feder Cooper and Katherine Lee and Matthew Jagielski and Milad Nasr and Arthur Conmy and Itay Yona and Eric Wallace and David Rolnick and Florian Tramèr},
      year={2024},
      eprint={2403.06634},
      archivePrefix={arXiv},
      primaryClass={cs.CR},
      url={https://arxiv.org/abs/2403.06634}, 
}

@misc{chen2018unsupervisedmultilingualwordembeddings,
      title={Unsupervised Multilingual Word Embeddings}, 
      author={Xilun Chen and Claire Cardie},
      year={2018},
      eprint={1808.08933},
      archivePrefix={arXiv},
      primaryClass={cs.CL},
      url={https://arxiv.org/abs/1808.08933}, 
}

@misc{chen2020graphoptimaltransportcrossdomain,
      title={Graph Optimal Transport for Cross-Domain Alignment}, 
      author={Liqun Chen and Zhe Gan and Yu Cheng and Linjie Li and Lawrence Carin and Jingjing Liu},
      year={2020},
      eprint={2006.14744},
      archivePrefix={arXiv},
      primaryClass={cs.CL},
      url={https://arxiv.org/abs/2006.14744}, 
}

@misc{conneau2018wordtranslationparalleldata,
      title={Word Translation Without Parallel Data}, 
      author={Alexis Conneau and Guillaume Lample and Marc'Aurelio Ranzato and Ludovic Denoyer and Hervé Jégou},
      year={2018},
      eprint={1710.04087},
      archivePrefix={arXiv},
      primaryClass={cs.CL},
      url={https://arxiv.org/abs/1710.04087}, 
}

@misc{granite2024embedding,
  title={Granite Embedding Models},
  url={https://github.com/ibm-granite/granite-embedding-models/},
  author={Granite Embedding Team, IBM},
  month=dec,
  year={2024}
}

@misc{grave2018unsupervisedalignmentembeddingswasserstein,
      title={Unsupervised Alignment of Embeddings with Wasserstein Procrustes}, 
      author={Edouard Grave and Armand Joulin and Quentin Berthet},
      year={2018},
      eprint={1805.11222},
      archivePrefix={arXiv},
      primaryClass={cs.LG},
      url={https://arxiv.org/abs/1805.11222}, 
}

@misc{huang2018multimodalunsupervisedimagetoimagetranslation,
      title={Multimodal Unsupervised Image-to-Image Translation}, 
      author={Xun Huang and Ming-Yu Liu and Serge Belongie and Jan Kautz},
      year={2018},
      eprint={1804.04732},
      archivePrefix={arXiv},
      primaryClass={cs.CV},
      url={https://arxiv.org/abs/1804.04732}, 
}

@misc{huh2024platonicrepresentationhypothesis,
      title={The Platonic Representation Hypothesis}, 
      author={Minyoung Huh and Brian Cheung and Tongzhou Wang and Phillip Isola},
      year={2024},
      eprint={2405.07987},
      archivePrefix={arXiv},
      primaryClass={cs.LG},
      url={https://arxiv.org/abs/2405.07987}, 
}

@misc{lample2018unsupervisedmachinetranslationusing,
      title={Unsupervised Machine Translation Using Monolingual Corpora Only}, 
      author={Guillaume Lample and Alexis Conneau and Ludovic Denoyer and Marc'Aurelio Ranzato},
      year={2018},
      eprint={1711.00043},
      archivePrefix={arXiv},
      primaryClass={cs.CL},
      url={https://arxiv.org/abs/1711.00043}, 
}

@misc{liu2018unsupervisedimagetoimagetranslationnetworks,
      title={Unsupervised Image-to-Image Translation Networks}, 
      author={Ming-Yu Liu and Thomas Breuel and Jan Kautz},
      year={2018},
      eprint={1703.00848},
      archivePrefix={arXiv},
      primaryClass={cs.CV},
      url={https://arxiv.org/abs/1703.00848}, 
}

@misc{maiorca2024latentspacetranslationsemantic,
      title={Latent Space Translation via Semantic Alignment}, 
      author={Valentino Maiorca and Luca Moschella and Antonio Norelli and Marco Fumero and Francesco Locatello and Emanuele Rodolà},
      year={2024},
      eprint={2311.00664},
      archivePrefix={arXiv},
      primaryClass={cs.LG},
      url={https://arxiv.org/abs/2311.00664}, 
}

@misc{moayeri2023texttoconceptandbackcrossmodel,
      title={Text-To-Concept (and Back) via Cross-Model Alignment}, 
      author={Mazda Moayeri and Keivan Rezaei and Maziar Sanjabi and Soheil Feizi},
      year={2023},
      eprint={2305.06386},
      archivePrefix={arXiv},
      primaryClass={cs.CV},
      url={https://arxiv.org/abs/2305.06386}, 
}

@misc{moschella2023relativerepresentationsenablezeroshot,
      title={Relative representations enable zero-shot latent space communication}, 
      author={Luca Moschella and Valentino Maiorca and Marco Fumero and Antonio Norelli and Francesco Locatello and Emanuele Rodolà},
      year={2023},
      eprint={2209.15430},
      archivePrefix={arXiv},
      primaryClass={cs.LG},
      url={https://arxiv.org/abs/2209.15430}, 
}

@misc{mrksic2016counterfittingwordvectorslinguistic,
      title={Counter-fitting Word Vectors to Linguistic Constraints}, 
      author={Nikola Mrksic and Diarmuid Ó Séaghdha and Blaise Thomson and Milica Gasic and Lina Rojas-Barahona and Pei-Hao Su and David Vandyke and Tsung-Hsien Wen and Steve Young},
      year={2016},
      eprint={1603.00892},
      archivePrefix={arXiv},
      primaryClass={cs.CL},
      url={https://arxiv.org/abs/1603.00892}, 
}

@misc{norelli2023asifcoupleddataturns,
      title={ASIF: Coupled Data Turns Unimodal Models to Multimodal Without Training}, 
      author={Antonio Norelli and Marco Fumero and Valentino Maiorca and Luca Moschella and Emanuele Rodolà and Francesco Locatello},
      year={2023},
      eprint={2210.01738},
      archivePrefix={arXiv},
      primaryClass={cs.LG},
      url={https://arxiv.org/abs/2210.01738}, 
}

@inproceedings{peyre2016gromov,
  title     = {Gromov-Wasserstein Averaging of Kernel and Distance Matrices},
  author    = {Gabriel Peyré and Marco Cuturi and Justin Solomon},
  booktitle = {Proceedings of the 33rd International Conference on Machine Learning (ICML)},
  year      = {2016},
  volume    = {48},
  series    = {JMLR: Workshop and Conference Proceedings},
  address   = {New York, NY, USA},
  publisher = {JMLR},
}

@inproceedings{raviknight2011deciphering,
    title = "Deciphering Foreign Language",
    author = "Ravi, Sujith  and
      Knight, Kevin",
    editor = "Lin, Dekang  and
      Matsumoto, Yuji  and
      Mihalcea, Rada",
    booktitle = "Proceedings of the 49th Annual Meeting of the Association for Computational Linguistics: Human Language Technologies",
    month = jun,
    year = "2011",
    address = "Portland, Oregon, USA",
    publisher = "Association for Computational Linguistics",
    url = "https://aclanthology.org/P11-1002/",
    pages = "12--21"
}

@misc{zhu2020unpairedimagetoimagetranslationusing,
      title={Unpaired Image-to-Image Translation using Cycle-Consistent Adversarial Networks}, 
      author={Jun-Yan Zhu and Taesung Park and Phillip Isola and Alexei A. Efros},
      year={2020},
      eprint={1703.10593},
      archivePrefix={arXiv},
      primaryClass={cs.CV},
      url={https://arxiv.org/abs/1703.10593}, 
}

@inproceedings{xing2015normalized,
    title = "Normalized Word Embedding and Orthogonal Transform for Bilingual Word Translation",
    author = "Xing, Chao  and
      Wang, Dong  and
      Liu, Chao  and
      Lin, Yiye",
    editor = "Mihalcea, Rada  and
      Chai, Joyce  and
      Sarkar, Anoop",
    booktitle = "Proceedings of the 2015 Conference of the North {A}merican Chapter of the Association for Computational Linguistics: Human Language Technologies",
    month = may,
    year = "2015",
    address = "Denver, Colorado",
    publisher = "Association for Computational Linguistics",
    url = "https://aclanthology.org/N15-1104/",
    doi = "10.3115/v1/N15-1104",
    pages = "1006--1011"
}

@misc{
yoon2025embeddingconverter,
title={Embedding-Converter: A Unified Framework for Cross-Model Embedding Transformation},
author={Jinsung Yoon and Sercan O Arik},
year={2025},
url={https://openreview.net/forum?id=ga9PAnFsAt}
}

@misc{dong2024dreamllmsynergisticmultimodalcomprehension,
      title={DreamLLM: Synergistic Multimodal Comprehension and Creation}, 
      author={Runpei Dong and Chunrui Han and Yuang Peng and Zekun Qi and Zheng Ge and Jinrong Yang and Liang Zhao and Jianjian Sun and Hongyu Zhou and Haoran Wei and Xiangwen Kong and Xiangyu Zhang and Kaisheng Ma and Li Yi},
      year={2024},
      eprint={2309.11499},
      archivePrefix={arXiv},
      primaryClass={cs.CV},
      url={https://arxiv.org/abs/2309.11499}, 
}

@article{heimerl2019embComp, title={embComp: Visual Interactive Comparison of Vector Embeddings}, volume={28}, ISSN={2160-9306}, url={http://dx.doi.org/10.1109/TVCG.2020.3045918}, DOI={10.1109/tvcg.2020.3045918}, number={8}, journal={IEEE Transactions on Visualization and Computer Graphics}, publisher={Institute of Electrical and Electronics Engineers (IEEE)}, author={Heimerl, Florian and Kralj, Christoph and Moller, Torsten and Gleicher, Michael}, year={2022}, month=aug, pages={2953–2969} }

@inproceedings{koh2023fromage,
    title={Grounding language models to images for multimodal inputs and outputs},
    author={Koh, Jing Yu and Salakhutdinov, Ruslan and Fried, Daniel},
    booktitle={International Conference on Machine Learning},
    pages={17283--17300},
    year={2023},
    organization={PMLR}
}

@misc{kornblith2019similarity,
      title={Similarity of Neural Network Representations Revisited}, 
      author={Simon Kornblith and Mohammad Norouzi and Honglak Lee and Geoffrey Hinton},
      year={2019},
      eprint={1905.00414},
      archivePrefix={arXiv},
      primaryClass={cs.LG},
      url={https://arxiv.org/abs/1905.00414}, 
}

@article{laakso2000content,
  title={Content and cluster analysis: Assessing representational similarity in neural systems},
  author={Laakso, Aarre and Cottrell, Garrison},
  journal={Philosophical Psychology},
  volume={13},
  number={1},
  pages={47--76},
  year={2000},
  doi={10.1080/09515080050002726}
}

@misc{li2016convergentlearningdifferentneural,
      title={Convergent Learning: Do different neural networks learn the same representations?}, 
      author={Yixuan Li and Jason Yosinski and Jeff Clune and Hod Lipson and John Hopcroft},
      year={2016},
      eprint={1511.07543},
      archivePrefix={arXiv},
      primaryClass={cs.LG},
      url={https://arxiv.org/abs/1511.07543}, 
}

@misc{liang2022mindgapunderstandingmodality,
      title={Mind the Gap: Understanding the Modality Gap in Multi-modal Contrastive Representation Learning}, 
      author={Weixin Liang and Yuhui Zhang and Yongchan Kwon and Serena Yeung and James Zou},
      year={2022},
      eprint={2203.02053},
      archivePrefix={arXiv},
      primaryClass={cs.CL},
      url={https://arxiv.org/abs/2203.02053}, 
}

@misc{li2023gte,
      title={Towards General Text Embeddings with Multi-stage Contrastive Learning}, 
      author={Zehan Li and Xin Zhang and Yanzhao Zhang and Dingkun Long and Pengjun Xie and Meishan Zhang},
      year={2023},
      eprint={2308.03281},
      archivePrefix={arXiv},
      primaryClass={cs.CL},
      url={https://arxiv.org/abs/2308.03281}, 
}

@misc{liu2023visualinstructiontuning,
      title={Visual Instruction Tuning}, 
      author={Haotian Liu and Chunyuan Li and Qingyang Wu and Yong Jae Lee},
      year={2023},
      eprint={2304.08485},
      archivePrefix={arXiv},
      primaryClass={cs.CV},
      url={https://arxiv.org/abs/2304.08485}, 
}

@misc{mokady2021clipcap,
      title={ClipCap: CLIP Prefix for Image Captioning}, 
      author={Ron Mokady and Amir Hertz and Amit H. Bermano},
      year={2021},
      eprint={2111.09734},
      archivePrefix={arXiv},
      primaryClass={cs.CV},
      url={https://arxiv.org/abs/2111.09734}, 
}

@misc{morcos2018insightsrepresentationalsimilarityneural,
      title={Insights on representational similarity in neural networks with canonical correlation}, 
      author={Ari S. Morcos and Maithra Raghu and Samy Bengio},
      year={2018},
      eprint={1806.05759},
      archivePrefix={arXiv},
      primaryClass={stat.ML},
      url={https://arxiv.org/abs/1806.05759}, 
}

@misc{morris2023textembeddingsrevealalmost,
      title={Text Embeddings Reveal (Almost) As Much As Text}, 
      author={John X. Morris and Volodymyr Kuleshov and Vitaly Shmatikov and Alexander M. Rush},
      year={2023},
      eprint={2310.06816},
      archivePrefix={arXiv},
      primaryClass={cs.CL},
      url={https://arxiv.org/abs/2310.06816}, 
}

@misc{ni2021gtr,
      title={Large Dual Encoders Are Generalizable Retrievers}, 
      author={Jianmo Ni and Chen Qu and Jing Lu and Zhuyun Dai and Gustavo Hernández Ábrego and Ji Ma and Vincent Y. Zhao and Yi Luan and Keith B. Hall and Ming-Wei Chang and Yinfei Yang},
      year={2021},
      eprint={2112.07899},
      archivePrefix={arXiv},
      primaryClass={cs.IR},
      url={https://arxiv.org/abs/2112.07899}, 
}

@misc{raghu2017svccasingularvectorcanonical,
      title={SVCCA: Singular Vector Canonical Correlation Analysis for Deep Learning Dynamics and Interpretability}, 
      author={Maithra Raghu and Justin Gilmer and Jason Yosinski and Jascha Sohl-Dickstein},
      year={2017},
      eprint={1706.05806},
      archivePrefix={arXiv},
      primaryClass={stat.ML},
      url={https://arxiv.org/abs/1706.05806}, 
}

@misc{schnaus2025itsblindmatchvisionlanguage,
      title={It's a (Blind) Match! Towards Vision-Language Correspondence without Parallel Data}, 
      author={Dominik Schnaus and Nikita Araslanov and Daniel Cremers},
      year={2025},
      eprint={2503.24129},
      archivePrefix={arXiv},
      primaryClass={cs.CV},
      url={https://arxiv.org/abs/2503.24129}, 
}

@misc{song2023bridgegapmodalitiescomprehensive,
      title={How to Bridge the Gap between Modalities: A Comprehensive Survey on Multimodal Large Language Model}, 
      author={Shezheng Song and Xiaopeng Li and Shasha Li and Shan Zhao and Jie Yu and Jun Ma and Xiaoguang Mao and Weimin Zhang},
      year={2023},
      eprint={2311.07594},
      archivePrefix={arXiv},
      primaryClass={cs.CL},
      url={https://arxiv.org/abs/2311.07594}, 
}

@misc{radford2021clip,
      title={Learning Transferable Visual Models From Natural Language Supervision}, 
      author={Alec Radford and Jong Wook Kim and Chris Hallacy and Aditya Ramesh and Gabriel Goh and Sandhini Agarwal and Girish Sastry and Amanda Askell and Pamela Mishkin and Jack Clark and Gretchen Krueger and Ilya Sutskever},
      year={2021},
      eprint={2103.00020},
      archivePrefix={arXiv},
      primaryClass={cs.CV},
      url={https://arxiv.org/abs/2103.00020}, 
}

@misc{wang2024e5,
      title={Text Embeddings by Weakly-Supervised Contrastive Pre-training}, 
      author={Liang Wang and Nan Yang and Xiaolong Huang and Binxing Jiao and Linjun Yang and Daxin Jiang and Rangan Majumder and Furu Wei},
      year={2024},
      eprint={2212.03533},
      archivePrefix={arXiv},
      primaryClass={cs.CL},
      url={https://arxiv.org/abs/2212.03533}, 
}

@misc{wang2023visionllm,
      title={VisionLLM: Large Language Model is also an Open-Ended Decoder for Vision-Centric Tasks}, 
      author={Wenhai Wang and Zhe Chen and Xiaokang Chen and Jiannan Wu and Xizhou Zhu and Gang Zeng and Ping Luo and Tong Lu and Jie Zhou and Yu Qiao and Jifeng Dai},
      year={2023},
      eprint={2305.11175},
      archivePrefix={arXiv},
      primaryClass={cs.CV},
      url={https://arxiv.org/abs/2305.11175}, 
}

@misc{yang2018unsupervisedneuralmachinetranslation,
      title={Unsupervised Neural Machine Translation with Weight Sharing}, 
      author={Zhen Yang and Wei Chen and Feng Wang and Bo Xu},
      year={2018},
      eprint={1804.09057},
      archivePrefix={arXiv},
      primaryClass={cs.CL},
      url={https://arxiv.org/abs/1804.09057}, 
}

@article{kwiatkowski-etal-2019-natural,
    title = "Natural Questions: A Benchmark for Question Answering Research",
    author = "Kwiatkowski, Tom  and
      Palomaki, Jennimaria  and
      Redfield, Olivia  and
      Collins, Michael  and
      Parikh, Ankur  and
      Alberti, Chris  and
      Epstein, Danielle  and
      Polosukhin, Illia  and
      Devlin, Jacob  and
      Lee, Kenton  and
      Toutanova, Kristina  and
      Jones, Llion  and
      Kelcey, Matthew  and
      Chang, Ming-Wei  and
      Dai, Andrew M.  and
      Uszkoreit, Jakob  and
      Le, Quoc  and
      Petrov, Slav",
    editor = "Lee, Lillian  and
      Johnson, Mark  and
      Roark, Brian  and
      Nenkova, Ani",
    journal = "Transactions of the Association for Computational Linguistics",
    volume = "7",
    year = "2019",
    address = "Cambridge, MA",
    publisher = "MIT Press",
    url = "https://aclanthology.org/Q19-1026",
    doi = "10.1162/tacl_a_00276",
    pages = "452--466",
}

@misc{girdhar2023imagebind,
      title={ImageBind: One Embedding Space To Bind Them All}, 
      author={Rohit Girdhar and Alaaeldin El-Nouby and Zhuang Liu and Mannat Singh and Kalyan Vasudev Alwala and Armand Joulin and Ishan Misra},
      year={2023},
      eprint={2305.05665},
      archivePrefix={arXiv},
      primaryClass={cs.CV},
      url={https://arxiv.org/abs/2305.05665}, 
}

@misc{li2023sentenceembeddingleaksinformation,
      title={Sentence Embedding Leaks More Information than You Expect: Generative Embedding Inversion Attack to Recover the Whole Sentence}, 
      author={Haoran Li and Mingshi Xu and Yangqiu Song},
      year={2023},
      eprint={2305.03010},
      archivePrefix={arXiv},
      primaryClass={cs.CL},
      url={https://arxiv.org/abs/2305.03010}, 
}

@misc{morris2023languagemodelinversion,
      title={Language Model Inversion}, 
      author={John X. Morris and Wenting Zhao and Justin T. Chiu and Vitaly Shmatikov and Alexander M. Rush},
      year={2023},
      eprint={2311.13647},
      archivePrefix={arXiv},
      primaryClass={cs.CL},
      url={https://arxiv.org/abs/2311.13647}, 
}

@misc{song2020informationleakageembeddingmodels,
      title={Information Leakage in Embedding Models}, 
      author={Congzheng Song and Ananth Raghunathan},
      year={2020},
      eprint={2004.00053},
      archivePrefix={arXiv},
      primaryClass={cs.LG},
      url={https://arxiv.org/abs/2004.00053}, 
}

@misc{wang2018understandinglearningrepresentationsextent,
      title={Towards Understanding Learning Representations: To What Extent Do Different Neural Networks Learn the Same Representation}, 
      author={Liwei Wang and Lunjia Hu and Jiayuan Gu and Yue Wu and Zhiqiang Hu and Kun He and John Hopcroft},
      year={2018},
      eprint={1810.11750},
      archivePrefix={arXiv},
      primaryClass={cs.LG},
      url={https://arxiv.org/abs/1810.11750}, 
}

@misc{yamagiwa2023discoveringuniversalgeometryembeddings,
      title={Discovering Universal Geometry in Embeddings with ICA}, 
      author={Hiroaki Yamagiwa and Momose Oyama and Hidetoshi Shimodaira},
      year={2023},
      eprint={2305.13175},
      archivePrefix={arXiv},
      primaryClass={cs.CL},
      url={https://arxiv.org/abs/2305.13175}, 
}

@misc{zhang2025stella,
      title={Jasper and Stella: distillation of SOTA embedding models}, 
      author={Dun Zhang and Jiacheng Li and Ziyang Zeng and Fulong Wang},
      year={2025},
      eprint={2412.19048},
      archivePrefix={arXiv},
      primaryClass={cs.IR},
      url={https://arxiv.org/abs/2412.19048}, 
}

@misc{zhang2024extractingpromptsinvertingllm,
      title={Extracting Prompts by Inverting LLM Outputs}, 
      author={Collin Zhang and John X. Morris and Vitaly Shmatikov},
      year={2024},
      eprint={2405.15012},
      archivePrefix={arXiv},
      primaryClass={cs.CL},
      url={https://arxiv.org/abs/2405.15012}, 
}

@misc{zhang2025universalzeroshotembeddinginversion,
      title={Universal Zero-shot Embedding Inversion}, 
      author={Collin Zhang and John X. Morris and Vitaly Shmatikov},
      year={2025},
      eprint={2504.00147},
      archivePrefix={arXiv},
      primaryClass={cs.CL},
      url={https://arxiv.org/abs/2504.00147}, 
}

@article{10.1145/3422622,
author = {Goodfellow, Ian and Pouget-Abadie, Jean and Mirza, Mehdi and Xu, Bing and Warde-Farley, David and Ozair, Sherjil and Courville, Aaron and Bengio, Yoshua},
title = {Generative adversarial networks},
year = {2020},
issue_date = {November 2020},
publisher = {Association for Computing Machinery},
address = {New York, NY, USA},
volume = {63},
number = {11},
issn = {0001-0782},
url = {https://doi.org/10.1145/3422622},
doi = {10.1145/3422622},
journal = {Commun. ACM},
month = oct,
pages = {139–144},
numpages = {6}
}

@inproceedings{antypas-etal-2024-multilingual,
    title = "Multilingual Topic Classification in {X}: Dataset and Analysis",
    author = "Antypas, Dimosthenis  and
      Ushio, Asahi  and
      Barbieri, Francesco  and
      Camacho-Collados, Jose",
    editor = "Al-Onaizan, Yaser  and
      Bansal, Mohit  and
      Chen, Yun-Nung",
    booktitle = "Proceedings of the 2024 Conference on Empirical Methods in Natural Language Processing",
    month = nov,
    year = "2024",
    address = "Miami, Florida, USA",
    publisher = "Association for Computational Linguistics",
    url = "https://aclanthology.org/2024.emnlp-main.1123/",
    doi = "10.18653/v1/2024.emnlp-main.1123",
    pages = "20136--20152"
}

@inproceedings{lehman-etal-2021-bert,
    title = "Does {BERT} Pretrained on Clinical Notes Reveal Sensitive Data?",
    author = "Lehman, Eric  and
      Jain, Sarthak  and
      Pichotta, Karl  and
      Goldberg, Yoav  and
      Wallace, Byron",
    editor = "Toutanova, Kristina  and
      Rumshisky, Anna  and
      Zettlemoyer, Luke  and
      Hakkani-Tur, Dilek  and
      Beltagy, Iz  and
      Bethard, Steven  and
      Cotterell, Ryan  and
      Chakraborty, Tanmoy  and
      Zhou, Yichao",
    booktitle = "Proceedings of the 2021 Conference of the North American Chapter of the Association for Computational Linguistics: Human Language Technologies",
    month = jun,
    year = "2021",
    address = "Online",
    publisher = "Association for Computational Linguistics",
    url = "https://aclanthology.org/2021.naacl-main.73/",
    doi = "10.18653/v1/2021.naacl-main.73",
    pages = "946--959"
}

@article{johnson2016mimic,
  title={MIMIC-III, a freely accessible critical care database},
  author={Johnson, Alistair EW and Pollard, Tom J and Shen, Lu and Lehman, Li-wei H and Feng, Mengling and Ghassemi, Mohammad and Moody, Benjamin and Szolovits, Peter and Anthony Celi, Leo and Mark, Roger G},
  journal={Scientific data},
  volume={3},
  number={1},
  pages={1--9},
  year={2016},
  publisher={Nature Publishing Group}
}

@article{kraljevic2021multi,
  title={Multi-domain clinical natural language processing with MedCAT: the medical concept annotation toolkit},
  author={Kraljevic, Zeljko and Searle, Thomas and Shek, Anthony and Roguski, Lukasz and Noor, Kawsar and Bean, Daniel and Mascio, Aurelie and Zhu, Leilei and Folarin, Amos A and Roberts, Angus and others},
  journal={Artificial intelligence in medicine},
  volume={117},
  pages={102083},
  year={2021},
  publisher={Elsevier}
}

@inproceedings{10.1007/978-3-540-30115-8_22,
author = {Klimt, Bryan and Yang, Yiming},
title = {The enron corpus: a new dataset for email classification research},
year = {2004},
isbn = {3540231056},
publisher = {Springer-Verlag},
address = {Berlin, Heidelberg},
url = {https://doi.org/10.1007/978-3-540-30115-8_22},
doi = {10.1007/978-3-540-30115-8_22},
booktitle = {Proceedings of the 15th European Conference on Machine Learning},
pages = {217–226},
numpages = {10},
location = {Pisa, Italy},
series = {ECML'04}
}

@article{10.1145/3446374,
author = {Saxena, Divya and Cao, Jiannong},
title = {Generative Adversarial Networks (GANs): Challenges, Solutions, and Future Directions},
year = {2021},
issue_date = {April 2022},
publisher = {Association for Computing Machinery},
address = {New York, NY, USA},
volume = {54},
number = {3},
issn = {0360-0300},
url = {https://doi.org/10.1145/3446374},
doi = {10.1145/3446374},
journal = {ACM Comput. Surv.},
month = may,
articleno = {63},
numpages = {42}
}

@article{wolfram2025layers,
  title={Layers at similar depths generate similar activations across llm architectures},
  author={Wolfram, Christopher and Schein, Aaron},
  journal={arXiv preprint arXiv:2504.08775},
  year={2025}
}

@inproceedings{mistry2023comparative,
  title={A Comparative Study of Sentence Embedding Models for Assessing Semantic Variation},
  author={Mistry, Deven M and Minai, Ali A},
  booktitle={International Conference on Artificial Neural Networks},
  pages={1--12},
  year={2023}
}

@inproceedings{maniparambil2024vision,
  title={Do Vision and Language Encoders Represent the World Similarly?},
  author={Maniparambil, Mayug and Akshulakov, Raiymbek and Djilali, Yasser Abdelaziz Dahou and El Amine Seddik, Mohamed and Narayan, Sanath and Mangalam, Karttikeya and O'Connor, Noel E},
  booktitle={Proceedings of the IEEE/CVF Conference on Computer Vision and Pattern Recognition},
  pages={14334--14343},
  year={2024}
}

@article{tian2019latent,
  title={Latent translation: Crossing modalities by bridging generative models},
  author={Tian, Yingtao and Engel, Jesse},
  journal={arXiv preprint arXiv:1902.08261},
  year={2019}
}

@article{li-etal-2024-vision-language,
    title = "Do Vision and Language Models Share Concepts? A Vector Space Alignment Study",
    author = "Li, Jiaang  and
      Kementchedjhieva, Yova  and
      Fierro, Constanza  and
      S{\o}gaard, Anders",
    journal = "Transactions of the Association for Computational Linguistics",
    volume = "12",
    year = "2024",
    address = "Cambridge, MA",
    publisher = "MIT Press",
    url = "https://aclanthology.org/2024.tacl-1.68/",
    doi = "10.1162/tacl_a_00698",
    pages = "1232--1249"
}

@inproceedings{artetxe-etal-2018-robust,
    title = "A robust self-learning method for fully unsupervised cross-lingual mappings of word embeddings",
    author = "Artetxe, Mikel  and
      Labaka, Gorka  and
      Agirre, Eneko",
    editor = "Gurevych, Iryna  and
      Miyao, Yusuke",
    booktitle = "Proceedings of the 56th Annual Meeting of the Association for Computational Linguistics (Volume 1: Long Papers)",
    month = jul,
    year = "2018",
    address = "Melbourne, Australia",
    publisher = "Association for Computational Linguistics",
    url = "https://aclanthology.org/P18-1073/",
    doi = "10.18653/v1/P18-1073",
    pages = "789--798"
}

@inproceedings{joulin-etal-2018-loss,
    title = "Loss in Translation: Learning Bilingual Word Mapping with a Retrieval Criterion",
    author = "Joulin, Armand  and
      Bojanowski, Piotr  and
      Mikolov, Tomas  and
      J{\'e}gou, Herv{\'e}  and
      Grave, Edouard",
    editor = "Riloff, Ellen  and
      Chiang, David  and
      Hockenmaier, Julia  and
      Tsujii, Jun{'}ichi",
    booktitle = "Proceedings of the 2018 Conference on Empirical Methods in Natural Language Processing",
    month = oct,
    year = "2018",
    address = "Brussels, Belgium",
    publisher = "Association for Computational Linguistics",
    url = "https://aclanthology.org/D18-1330/",
    doi = "10.18653/v1/D18-1330",
    pages = "2979--2984"
}

@inproceedings{lin2014microsoft,
  title={Microsoft coco: Common objects in context},
  author={Lin, Tsung-Yi and Maire, Michael and Belongie, Serge and Hays, James and Perona, Pietro and Ramanan, Deva and Doll{\'a}r, Piotr and Zitnick, C Lawrence},
  booktitle={European conference on computer vision},
  pages={740--755},
  year={2014},
  organization={Springer}
}

@article{qwen3embedding,
  title={Qwen3 Embedding: Advancing Text Embedding and Reranking Through Foundation Models},
  author={Zhang, Yanzhao and Li, Mingxin and Long, Dingkun and Zhang, Xin and Lin, Huan and Yang, Baosong and Xie, Pengjun and Yang, An and Liu, Dayiheng and Lin, Junyang and Huang, Fei and Zhou, Jingren},
  journal={arXiv preprint arXiv:2506.05176},
  year={2025}
}


\newpage
\appendix
\section{Compute}
\label{sec:compute}
Our training and evaluation were conducted using diverse compute environments, including both local and cloud GPU clusters. Experiments were done on NVIDIA 2080Ti, L4, A40, and A100 GPUs, listed in order of increasing computational capacity.

For our final results, we trained 25 \atk{} models fully and 30 models partially (see \Cref{sec:stability}). The full models' training durations usually ranged from 1 to 7 days, depending on the specific GPU and model pair (which affected convergence rates). Partial convergence was stopped after 2 days. Due to the size of Qwen, our (qwen, gte) ablation was trained for 20 days on an A100.  Taking a conservative estimate of the average training time, this amounted to approximately 176 GPU days (24 models $\times$ 4 days / model + 30 models $\times$ 2 days / model + 1 (qwen, gte) $\times$ 20 days / model).

Evaluation procedures varied by model type:
\begin{itemize}
    \item The 10 main \atk{} models required $\sim$1 hour each for NQ, TweetTopic, and MIMIC evaluation (across GPU types), plus 30 minutes for attribute extraction on TweetTopic and MIMIC, and 1.5 hours for inversion and downstream LLM evaluation on Enron and TweetTopic. Naive baselines required $\sim$30 minutes each across all datasets.
    \item The 15 additional fully-trained models required 30 minutes each for NQ evaluation, with an extra 30 minutes for MS COCO evaluation of (clip, granite).
    \item Optimal transport baselines ran on CPU only, requiring $\sim$1 hour per dataset (three datasets for main models, one for others).
\end{itemize}

In total, our experiments consumed almost 176 GPU days for training and an additional 42 GPU hours for evaluation and analysis. An additional 45 CPU hours were required for optimal transport.

\section{Oracle-aided optimal transport baseline}
\label{sec:ot_baseline}
Let $u_i = M_1(d_i)$ and $v_i = M_2(d_i)$ denote embeddings of the same document $d_i$ from two different embedding models. In \Cref{sec:eval_geometry}, we solve the optimal assignment problem:
\[
\pi^* = \arg\min_{\pi} \sum_{i=1}^n \cos(u_i, v_{\pi(i)}),
\]
using four algorithms: Hungarian (linear sum assignment), Earth Mover's Distance (EMD), Sinkhorn, Gromov-Wasserstein. For the Gromov-Wasserstein algorithm, we try both the entropic and non-entropic variants with multiple hyperparameter configurations and select the best figure. Note that the optimal transport (OT) baseline computes matchings and transports between embeddings derived from the \textit{same underlying texts}, strongly favoring OT methods. Nevertheless, OT still struggles when embeddings originate from different model backbones.

Since the Hungarian algorithm produces a discrete matching, it is evaluated only using Top-1 Accuracy, while the other algorithms are evaluated across all metrics. For each experiment, the lowest-rank solver is reported in \Cref{tab:embedding_comparison} and \Cref{tab:CLIP_nq} (denoted by symbols in the final column). Evaluation metrics are defined as follows:

\begin{enumerate}
    \item \textbf{Top-1 Accuracy}: Fraction of embeddings correctly identified as closest pairs, calculated by either selecting the maximum transported mass per embedding or applying the Hungarian algorithm directly to the transport plan $P$. We report the higher accuracy between the two.
    \item \textbf{Mean Rank}: Average rank position of the correct embedding match $v_i$ when sorted by descending transported mass $P_{ij}$ from $u_i$:
    \[
    \text{rank}(v_i) = \text{position of } v_i \text{ among sorted } P_{ij}.
    \]
    \item \textbf{Mean Cosine Similarity}: Average cosine similarity between barycenters and true counterparts:
    \[
    v'_i = \frac{\sum_{j=1}^n P_{ij} v_j}{\sum_{j=1}^n P_{ij}},\quad
    \text{Similarity} = \frac{1}{n}\sum_{i=1}^{n}\cos(v'_i,v_i).
    \]
\end{enumerate}
\newpage
\section{Translating to and from Qwen}
\label{sec:qwen}
\begin{table}[!h]
    \scriptsize
    \centering
    \begin{tabular}{llrrrrrr}
        \toprule
        \multicolumn{2}{c}{} & \multicolumn{3}{c}{\atk{}} & \multicolumn{3}{c}{OT Baseline}\\
        \cmidrule(lr){3-5} \cmidrule(lr){6-8}
        $M_1$ & $M_2$ & $\cos(\cdot)$ $\uparrow$ & T-1 $\uparrow$ & Rank $\downarrow$& $\cos(\cdot)$ $\uparrow$ & T-1 $\uparrow$ & Rank $\downarrow$\\
        \midrule
        gte & qwen & \textbf{0.50 {\tiny (0.0)}} & \textbf{0.92} & \textbf{2.28 {\tiny (0.2)}} & 0.38 {\tiny (0.0)} & 0.00 & 425.28 {\tiny (1.1)}$^{\ddagger}$ \\
        qwen & gte & 0.84 {\tiny (0.0)} & \textbf{0.88} & \textbf{2.49 {\tiny (0.3)}} & \textbf{0.85 {\tiny (0.0)}} & 0.00 & 425.07 {\tiny (1.2)}$^{\ddagger}$ \\
        \bottomrule\\[-1.6ex]
    \end{tabular}
    \caption{Translations between GTE and Qwen embeddings trained on NQ and evaluated on a 65536 text subset of NQ (chunked in batches of size 1024). Rank varies from 1 to 1024, thus 512 corresponds to a random ordering. Since the embedding dimensionalities are different, only the Gromov-Wasserstein$^{\ddagger}$ OT baseline is run and the naive baseline does not apply. Bold denotes best value.}
    \label{tab:qwen_nq}
\end{table} 

As shown in \Cref{tab:qwen_nq}, \atk{} successfully translates between GTE and Qwen, significantly outperforming the optimal transport baseline in all metrics except qwen $\to$ gte cosine similarity, which we hypothesize may be due to the substantial performance gap between the models---indeed, Qwen differs from GTE in architecture (dense Qwen backbone), training methodology (unsupervised + model merging techniques), size (14$\times$ larger than the next largest model and 37$\times$ larger than GTE), context length, and recency. Given Qwen's size and computational cost, we only evaluated this representative pair. We leave further evaluation to future work.

\section{Text-image retrieval on MS COCO}
\label{sec:mscoco}

\begin{table}[!ht]
    \scriptsize
    \centering
    \begin{tabular}{lrrr}
        \toprule
        model& R@16 $\uparrow$ &$\cos(\cdot)$ $\uparrow$ & Rank $\downarrow$\\
        \midrule
        granite $\to$ clip&0.23&0.23 {\tiny (0.0)}&233.67 {\tiny (3.0)}\\
        clip (baseline) &0.75&0.30 {\tiny (0.0)}&23.20 {\tiny (0.8)}\\
        \bottomrule\\[-1.6ex]
    \end{tabular}
    \caption{Cross-model text-image retrieval on MS COCO: granite $\to$ clip \atk{} trained on NQ (unimodal) and evaluated on MS COCO's validation set. The rank metric varies from 1 to 5000, thus 2500 corresponds to a random ordering. Queries (captions) embedded with either Granite or CLIP. Documents (images) embedded with CLIP. Each caption has a unique image. Standard errors are shown in parentheses.}
    \label{tab:msmarco}
\end{table}

Our \atk{}s can ``stitch" modalities onto unimodal models by translating to a multimodal model. To test this, we evaluated cross-modal text-image retrieval on MS COCO's validation set (5000 examples) \cite{lin2014microsoft}, translating queries (captions) embedded with Granite to retrieve documents (images) embedded with CLIP using our unimodal granite $\to$ clip translator from \cref{sec:extracting_attributes}. Each caption has a unique image. We report Recall@16, cosine similarities, and Rank, with CLIP (for both documents and queries) as our baseline.

As \Cref{tab:msmarco} shows, translating Granite embeddings to CLIP enables non-negligible cross-model multimodal retrieval with \textbf{a unimodal model for queries}—despite zero multimodal training. Further evaluation of this paradigm with multimodal-specific training is a promising direction.

\section{Initialization robustness by model backbone}
\label{sec:stability}

GAN training is notoriously unstable to weight initialization~\cite{10.1145/3446374}. To measure our method's robustness, we trained fifteen e5 $\to$ gte (shared backbone) and e5 $\to$ gtr (cross-backbone) \atk{}s on the NQ dataset for a fixed 10 epochs.

For the translations between related models, \atk{} training was relatively stable across random seeds: 14 out of 15 seeds achieved at least 80\% top-1 accuracy within a fixed epoch budget, while the remaining run reached 72\%. In contrast, translation between unrelated models proved significantly less stable, with only 3 out of 15 runs achieving convergence (80\% top-1 accuracy). We leave improving the seed stability of our training regime as future, valuable work.

\section{Full out-of-distribution translation results}
\label{sec:full_ood}
We provide baseline numbers for the experiments shown in \Cref{tab:embeddings_ood}, by dataset.

\begin{table}[!ht]
    \scriptsize
    \centering
    \begin{tabular}{llrrrrrrrrr}
        \toprule
        \multicolumn{2}{c}{} & \multicolumn{3}{c}{\atk{}} & \multicolumn{3}{c}{Naïve Baseline} & \multicolumn{3}{c}{OT Baseline}\\
        \cmidrule(lr){3-5} \cmidrule(lr){6-8}  \cmidrule(lr){9-11} 
        $E_1$ & $E_2$ & $\cos(\cdot)$ $\uparrow$ & T-1 $\uparrow$ & Rank $\downarrow$ & $\cos(\cdot)$ $\uparrow$ & T-1 $\uparrow$ & Rank $\downarrow$ &  $\cos(\cdot)$ $\uparrow$ & T-1 $\uparrow$ & Rank $\downarrow$\\
        \midrule
        \multirow{4}{*}{gra.} & gtr & 0.74 {\tiny (0.0)} & 0.99 & 1.09 {\tiny (0.1)} & -0.04 {\tiny (0.0)} & 0.00 & 415.61 {\tiny (8.2)} & 0.71 {\tiny (0.0)} & 0.01 & 220.93 {\tiny (7.1)}$^{\ddagger}$ \\
        & gte & 0.85 {\tiny (0.0)} & 0.95 & 1.26 {\tiny (0.1)} & 0.00 {\tiny (0.0)} & 0.00 & 406.73 {\tiny (8.2)} & 0.87 {\tiny (0.0)} & 0.01 & 201.48 {\tiny (6.6)}$^{\ddagger}$ \\
        & ste. & 0.77 {\tiny (0.0)} & 0.96 & 1.11 {\tiny (0.0)} & 0.00 {\tiny (0.0)} & 0.00 & 417.27 {\tiny (8.2)} & 0.74 {\tiny (0.0)} & 0.00 & 239.36 {\tiny (6.7)}$^{\ddagger}$ \\
        & e5 & 0.83 {\tiny (0.0)} & 0.87 & 3.10 {\tiny (0.7)} & 0.02 {\tiny (0.0)} & 0.00 & 405.53 {\tiny (8.1)} & 0.87 {\tiny (0.0)} & 0.01 & 244.94 {\tiny (7.4)}$^{\ddagger}$ \\
        \midrule
        \multirow{4}{*}{gtr} & gra. & 0.79 {\tiny (0.0)} & 0.98 & 2.41 {\tiny (0.6)} & -0.04 {\tiny (0.0)} & 0.00 & 411.53 {\tiny (8.3)} & 0.57 {\tiny (0.0)} & 0.01 & 398.29 {\tiny (8.2)}$^\ddagger$ \\
        & gte & 0.85 {\tiny (0.0)} & 0.96 & 1.29 {\tiny (0.2)} & 0.04 {\tiny (0.0)} & 0.00 & 392.01 {\tiny (8.2)} & 0.86 {\tiny (0.0)} & 0.01 & 259.47 {\tiny (7.4)}$^{\ddagger}$ \\
        & ste. & 0.77 {\tiny (0.0)} & 0.96 & 1.10 {\tiny (0.0)} & 0.00 {\tiny (0.0)} & 0.00 & 394.69 {\tiny (8.3)} & 0.74 {\tiny (0.0)} & 0.00 & 294.58 {\tiny (7.4)}$^{\ddagger}$ \\
        & e5 & 0.80 {\tiny (0.0)} & 0.53 & 13.38 {\tiny (1.2)} & 0.03 {\tiny (0.0)} & 0.00 & 400.85 {\tiny (8.2)} & 0.87 {\tiny (0.0)} & 0.01 & 266.04 {\tiny (7.7)}$^{\ddagger}$ \\
        \midrule
        \multirow{4}{*}{gte} & gra. & 0.73 {\tiny (0.0)} & 0.94 & 1.33 {\tiny (0.1)} & 0.00 {\tiny (0.0)} & 0.00 & 408.81 {\tiny (8.3)} & 0.56 {\tiny (0.0)} & 0.01 & 398.16 {\tiny (8.2)}$^*$ \\
        & gtr & 0.71 {\tiny (0.0)} & 0.95 & 1.29 {\tiny (0.1)} & 0.04 {\tiny (0.0)} & 0.00 & 386.58 {\tiny (8.3)} & 0.71 {\tiny (0.0)} & 0.01 & 254.74 {\tiny (7.3)}$^{\ddagger}$ \\
        & ste. & 0.86 {\tiny (0.0)} & 1.00 & 1.00 {\tiny (0.0)} & 0.58 {\tiny (0.0)} & 1.00 & 1.00 {\tiny (0.0)} & 1.00 {\tiny (0.0)} & 1.00 & 1.00 {\tiny (0.0)}$^{*}$ \\
        & e5 & 0.83 {\tiny (0.0)} & 0.91 & 1.57 {\tiny (0.2)} & 0.68 {\tiny (0.0)} & 1.00 & 1.00 {\tiny (0.0)} & 1.00 {\tiny (0.0)} & 1.00 & 1.00 {\tiny (0.0)}$^{*}$ \\
        \midrule
        \multirow{4}{*}{ste.} & gra. & 0.79 {\tiny (0.0)} & 0.99 & 1.09 {\tiny (0.1)} & 0.00 {\tiny (0.0)} & 0.00 & 418.16 {\tiny (8.4)} & 0.57 {\tiny (0.0)} & 0.00 & 399.56 {\tiny (8.2)}$^\ddagger$ \\
        & gtr & 0.77 {\tiny (0.0)} & 1.00 & 1.00 {\tiny (0.0)} & 0.00 {\tiny (0.0)} & 0.00 & 393.07 {\tiny (8.1)} & 0.71 {\tiny (0.0)} & 0.00 & 294.65 {\tiny (7.4)}$^{\ddagger}$ \\
        & gte & 0.90 {\tiny (0.0)} & 1.00 & 1.00 {\tiny (0.0)} & 0.58 {\tiny (0.0)} & 1.00 & 1.00 {\tiny (0.0)} & 1.00 {\tiny (0.0)} & 1.00 & 1.00 {\tiny (0.0)}$^{*}$ \\
        & e5 & 0.85 {\tiny (0.0)} & 0.98 & 1.05 {\tiny (0.0)} & 0.37 {\tiny (0.0)} & 0.89 & 1.55 {\tiny (0.1)} & 1.00 {\tiny (0.0)} & 1.00 & 1.00 {\tiny (0.0)}$^{*}$ \\
        \midrule
        \multirow{4}{*}{e5} & gra. & 0.79 {\tiny (0.0)} & 0.98 & 1.08 {\tiny (0.0)} & 0.02 {\tiny (0.0)} & 0.00 & 405.75 {\tiny (8.3)} & 0.57 {\tiny (0.0)} & 0.01 & 398.34 {\tiny (8.2)}$^\ddagger$ \\
        & gtr & 0.67 {\tiny (0.0)} & 0.80 & 3.10 {\tiny (0.6)} & 0.03 {\tiny (0.0)} & 0.00 & 401.16 {\tiny (8.4)} & 0.71 {\tiny (0.0)} & 0.00 & 268.28 {\tiny (7.6)}$^{\ddagger}$ \\
        & gte & 0.87 {\tiny (0.0)} & 0.99 & 1.02 {\tiny (0.0)} & 0.68 {\tiny (0.0)} & 1.00 & 1.00 {\tiny (0.0)} & 1.00 {\tiny (0.0)} & 1.00 & 1.00 {\tiny (0.0)}$^{*}$ \\
        & ste. & 0.75 {\tiny (0.0)} & 0.98 & 1.06 {\tiny (0.0)} & 0.37 {\tiny (0.0)} & 1.00 & 1.00 {\tiny (0.0)} & 1.00 {\tiny (0.0)} & 1.00 & 1.00 {\tiny (0.0)}$^{*}$ \\
        \bottomrule\\[-1.6ex]
    \end{tabular}
    \caption{Out-of-distribution translations on TweetTopic (with baselines): \atk{} models trained on NQ and evaluated on the entire TweetTopic test set (800 tweets). The rank metric varies from 1 to 800, thus 400 corresponds to a random ordering. Standard errors are shown in parentheses. Symbols denote the lowest-rank solver: Earth Mover's Distance$^*$ and Gromov-Wasserstein$^\ddagger$}
    \label{tab:embedding_comparison_ood}
\end{table}

\begin{table}[!ht]
    \scriptsize
    \centering
    \begin{tabular}{llrrrrrrrrr}
        \toprule
        \multicolumn{2}{c}{} & \multicolumn{3}{c}{\atk{}} & \multicolumn{3}{c}{Naïve Baseline} & \multicolumn{3}{c}{OT Baseline}\\
        \cmidrule(lr){3-5} \cmidrule(lr){6-8}  \cmidrule(lr){9-11} 
        $E_1$ & $E_2$ & $\cos(\cdot)$ $\uparrow$ & T-1 $\uparrow$ & Rank $\downarrow$ & $\cos(\cdot)$ $\uparrow$ & T-1 $\uparrow$ & Rank $\downarrow$ & $\cos(\cdot)$ $\uparrow$ & T-1 $\uparrow$ & Rank $\downarrow$\\
        \midrule
        \multirow{4}{*}{gra.} & gtr & 0.74 {\tiny (0.0)} & 0.60 & 23.38 {\tiny (1.6)} & -0.02 {\tiny (0.0)} & 0.00 & 4010.00 {\tiny (25.8)} & 0.82 {\tiny (0.0)} & 0.00 & 3962.83 {\tiny (26.1)}$^\dagger$ \\
        & gte & 0.85 {\tiny (0.0)} & 0.08 & 346.21 {\tiny (7.8)} & 0.01 {\tiny (0.0)} & 0.00 & 3978.35 {\tiny (26.1)} & 0.92 {\tiny (0.0)} & 0.00 & 3808.18 {\tiny (25.9)}$^\dagger$ \\
        & ste. & 0.72 {\tiny (0.0)} & 0.13 & 242.23 {\tiny (6.1)} & -0.01 {\tiny (0.0)} & 0.00 & 3900.74 {\tiny (26.2)} & 0.86 {\tiny (0.0)} & 0.02 & 3780.44 {\tiny (26.0)}$^\dagger$ \\
        & e5 & 0.84 {\tiny (0.0)} & 0.12 & 361.06 {\tiny (8.7)} & 0.02 {\tiny (0.0)} & 0.00 & 4024.92 {\tiny (26.1)} & 0.93 {\tiny (0.0)} & 0.00 & 3937.63 {\tiny (26.2)}$^\dagger$ \\
        \midrule
        \multirow{4}{*}{gtr} & gra. & 0.78 {\tiny (0.0)} & 0.51 & 35.27 {\tiny (1.9)} & -0.02 {\tiny (0.0)} & 0.00 & 4023.67 {\tiny (26.1)} & 0.87 {\tiny (0.0)} & 0.00 & 3964.83 {\tiny (26.1)}$^\dagger$ \\
        & gte & 0.84 {\tiny (0.0)} & 0.12 & 279.56 {\tiny (6.9)} & 0.08 {\tiny (0.0)} & 0.00 & 4180.47 {\tiny (26.2)} & 0.87 {\tiny (0.0)} & 0.00 & 4088.97 {\tiny (26.2)}$^\ddagger$ \\
        & ste. & 0.72 {\tiny (0.0)} & 0.27 & 127.92 {\tiny (4.4)} & 0.00 {\tiny (0.0)} & 0.00 & 4296.04 {\tiny (26.1)} & 0.76 {\tiny (0.0)} & 0.00 & 4095.11 {\tiny (26.1)}$^\ddagger$ \\
        & e5 & 0.82 {\tiny (0.0)} & 0.01 & 1413.80 {\tiny (18.3)} & 0.09 {\tiny (0.0)} & 0.00 & 4064.47 {\tiny (26.2)} & 0.93 {\tiny (0.0)} & 0.00 & 4010.13 {\tiny (26.1)}$^\dagger$ \\
        \midrule
        \multirow{4}{*}{gte} & gra. & 0.73 {\tiny (0.0)} & 0.09 & 342.15 {\tiny (7.8)} & 0.01 {\tiny (0.0)} & 0.00 & 3946.19 {\tiny (25.8)} & 0.87 {\tiny (0.0)} & 0.00 & 3802.92 {\tiny (25.9)}$^\dagger$ \\
        & gtr & 0.69 {\tiny (0.0)} & 0.12 & 256.63 {\tiny (6.4)} & 0.08 {\tiny (0.0)} & 0.00 & 4229.90 {\tiny (26.2)} & 0.69 {\tiny (0.0)} & 0.00 & 4094.02 {\tiny (26.1)}$^\ddagger$ \\
        & ste. & 0.85 {\tiny (0.0)} & 1.00 & 1.00 {\tiny (0.0)} & 0.56 {\tiny (0.0)} & 1.00 & 1.00 {\tiny (0.0)} & 1.00 {\tiny (0.0)} & 1.00 & 1.00 {\tiny (0.0)}$^*$ \\
        & e5 & 0.86 {\tiny (0.0)} & 0.54 & 17.71 {\tiny (0.9)} & 0.69 {\tiny (0.0)} & 0.98 & 1.04 {\tiny (0.0)} & 1.00 {\tiny (0.0)} & 1.00 & 1.00 {\tiny (0.0)}$^*$ \\
        \midrule
        \multirow{4}{*}{ste.} & gra. & 0.77 {\tiny (0.0)} & 0.14 & 221.95 {\tiny (5.9)} & -0.01 {\tiny (0.0)} & 0.00 & 3951.42 {\tiny (25.9)} & 0.87 {\tiny (0.0)} & 0.01 & 3776.52 {\tiny (26.0)}$^\dagger$ \\
        & gtr & 0.75 {\tiny (0.0)} & 0.56 & 17.70 {\tiny (1.0)} & 0.00 {\tiny (0.0)} & 0.00 & 4339.83 {\tiny (26.2)} & 0.70 {\tiny (0.0)} & 0.00 & 4093.61 {\tiny (26.1)}$^\ddagger$ \\
        & gte & 0.91 {\tiny (0.0)} & 1.00 & 1.00 {\tiny (0.0)} & 0.56 {\tiny (0.0)} & 1.00 & 1.00 {\tiny (0.0)} & 1.00 {\tiny (0.0)} & 1.00 & 1.00 {\tiny (0.0)}$^*$ \\
        & e5 & 0.85 {\tiny (0.0)} & 0.51 & 26.33 {\tiny (1.2)} & 0.35 {\tiny (0.0)} & 0.59 & 12.68 {\tiny (0.6)} & 0.93 {\tiny (0.0)} & 1.00 & 1.00 {\tiny (0.0)}$^\dagger$ \\
        \midrule
        \multirow{4}{*}{e5} & gra. & 0.78 {\tiny (0.0)} & 0.21 & 151.09 {\tiny (4.6)} & 0.02 {\tiny (0.0)} & 0.00 & 4008.10 {\tiny (25.9)} & 0.87 {\tiny (0.0)} & 0.00 & 3932.58 {\tiny (26.2)}$^\dagger$ \\
        & gtr & 0.66 {\tiny (0.0)} & 0.01 & 1029.64 {\tiny (14.9)} & 0.09 {\tiny (0.0)} & 0.00 & 4032.85 {\tiny (26.2)} & 0.82 {\tiny (0.0)} & 0.00 & 4010.06 {\tiny (26.1)}$^\dagger$ \\
        & gte & 0.87 {\tiny (0.0)} & 0.60 & 32.59 {\tiny (2.6)} & 0.69 {\tiny (0.0)} & 0.98 & 1.09 {\tiny (0.0)} & 1.00 {\tiny (0.0)} & 1.00 & 1.00 {\tiny (0.0)}$^*$ \\
        & ste. & 0.75 {\tiny (0.0)} & 0.46 & 32.12 {\tiny (1.4)} & 0.35 {\tiny (0.0)} & 0.86 & 2.49 {\tiny (0.1)} & 0.86 {\tiny (0.0)} & 1.00 & 1.01 {\tiny (0.0)}$^\dagger$ \\
        \bottomrule\\[-1.6ex]
    \end{tabular}
    \caption{Out-of-distribution translations on MIMIC (with baselines): \atk{} models trained on NQ and evaluated on an 8192-record subset of MIMIC. The rank metric varies from 1 to 8192, thus 4096 corresponds to a random ordering. Standard errors are shown in parentheses. Symbols denote the lowest-rank solver: Earth Mover's Distance$^*$, Sinkhorn$^\dagger$ and Gromov-Wasserstein$^\ddagger$}
    \label{tab:embedding_comparison_ood_mimic}
\end{table}

\newpage

\section{Zero-shot inversion on TweetTopic}
\label{sec:zs-tt}
\begin{figure}[!ht]
    \centering
    \includegraphics[width=0.5\linewidth]{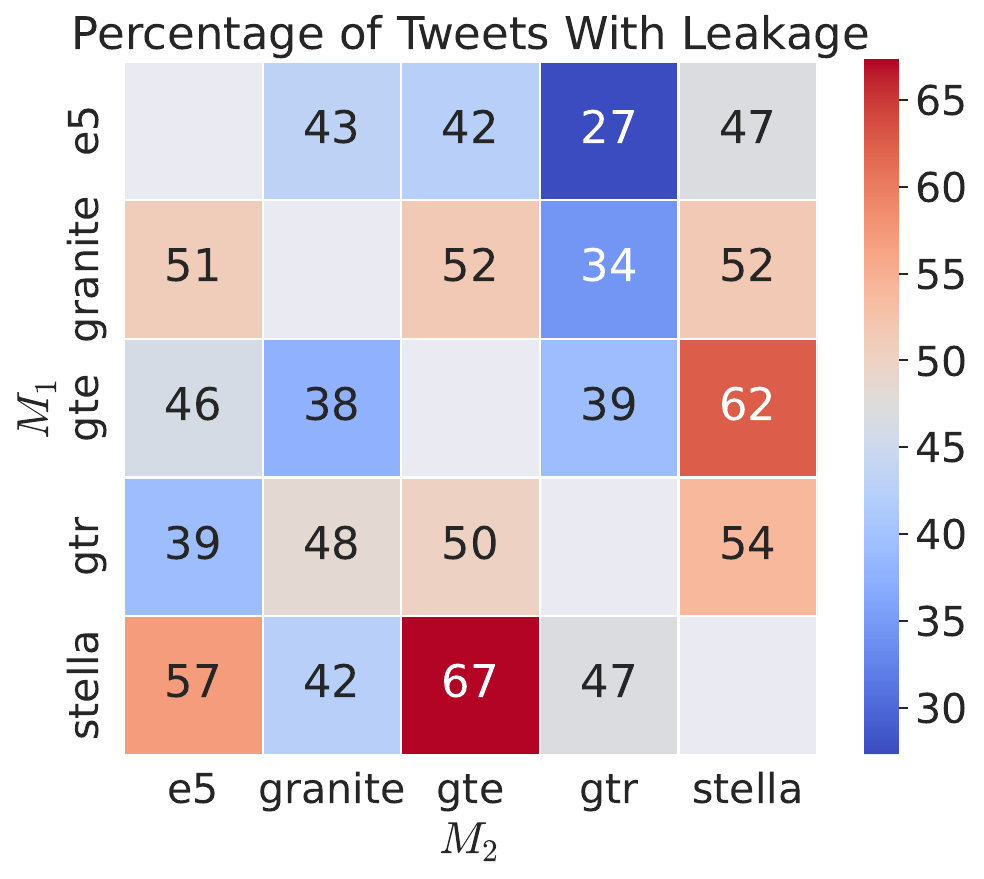}
    \caption{Leakage of information via inversion. Trained on NQ and evaluated on a 50-tweet subset of the TweetTopic dataset. Cells denote judge accuracy.}
    \label{fig:tweettopic_heatmap}
\end{figure}
We replicate \Cref{fig:enron_heatmap} on the TweetTopic dataset in \Cref{fig:tweettopic_heatmap}, extracting information given \textit{only} translated embeddings of tweets. We achieve non-negligible leakage for all model pairs.

\section{Prompt for measuring information extraction}
\label{sec:prompt}

\Cref{fig:enron_heatmap,fig:tweettopic_heatmap} utilize an LLM judge to determine whether the approximate inversion (using~\cite{zhang2025universalzeroshotembeddinginversion}) of a translated email embedding leaks information about the original email.   \Cref{fig:prompt} shows our prompt to the GPT-4o judge:
\begin{figure}[ht]
  \centering
    \begin{callout}
    \textbf{Original email:} \{ground\_truth\}
    
    \textbf{Reconstructed email:} \{generation\}.

    Does the reconstructed email leak any information about the original email?
    
    Answer with only `yes' or `no'.
    \end{callout}
  \caption{The prompt given to the LLM judge.}
  \label{fig:prompt}
\end{figure}



\end{document}